\documentclass[]{amap}

\usepackage[toc,page,header]{appendix}
\usepackage{minitoc}
\usepackage{solarized-light}
\usepackage{array}
\usepackage{siunitx,array}
\usepackage{makecell}
\usepackage{framed}

\usepackage{bbding}
\usepackage{amsmath}
\usepackage{amsfonts}
\usepackage[colorinlistoftodos]{todonotes}
\usepackage{longtable}
\usepackage{hhline}
\usepackage{fancyvrb}
\usepackage{float}
\usepackage{fvextra}
\usepackage{CJKutf8}
\usepackage{multicol}
\usepackage{tablefootnote}
\usepackage{threeparttable}
\usepackage{tabularx}
\usepackage{mdframed}
\usepackage[usestackEOL]{stackengine}
\usepackage[numbers]{natbib}
\newcommand{\commentout}[1]{}
\renewcommand{\paragraph}[1]
{\noindent\textbf{#1.}\hspace*{1em}}
\usepackage{enumitem}
\setlist[itemize]{leftmargin=15pt}
\usepackage{adjustbox}
\usepackage{arydshln}
\usepackage{pifont}

\RequirePackage{xspace}
\makeatletter
\DeclareRobustCommand\onedot{\futurelet\@let@token\@onedot}
\def\@onedot{\ifx\@let@token.\else.\null\fi\xspace}

\def\eg{\emph{e.g}\onedot}

\makeatother

\title{ABot-PhysWorld: Interactive World Foundation Model for
Robotic Manipulation with Physics Alignment}

\author{AMAP CV Lab}
\vspace{-11pt}

\contribution{See \hyperref[sec:contributions]{Contributions} section for a full author list.}

\abstract{
Video-based world models offer a powerful paradigm for embodied simulation and planning, yet state-of-the-art models often generate physically implausible manipulations—such as object penetration and anti-gravity motion—due to training on generic visual data and likelihood-based objectives that ignore physical laws.
We present \textbf{ABot-PhysWorld}, a 14B Diffusion Transformer model that generates visually realistic, physically plausible, and action-controllable videos. Built on a curated dataset of three million manipulation clips with physics-aware annotation, it uses a novel DPO-based post-training framework with decoupled discriminators to suppress unphysical behaviors while preserving visual quality. A parallel context block enables precise spatial action injection for cross-embodiment control.
To better evaluate generalization, we introduce \textbf{EZSbench}, the first training-independent embodied zero-shot benchmark combining real and synthetic unseen robot-task-scene combinations. It employs a decoupled protocol to separately assess physical realism and action alignment.
ABot-PhysWorld achieves new state-of-the-art performance on PBench and EZSbench, surpassing Veo~3.1 and Sora~v2~Pro in physical plausibility and trajectory consistency. We will release EZSbench to promote standardized evaluation in embodied video generation.
\bigskip

\textbf{Date:} March 20, 2026

\textbf{Correspondence:} xumu.xm@alibaba-inc.com

\textbf{Project Page:} \url{https://github.com/amap-cvlab/ABot-PhysWorld}
}

\begin{document}

\maketitle
\vspace{-4pt}

\begin{figure}[!h]
    \centering
    % \vspace{-10pt}
\includegraphics[width=0.8\linewidth]{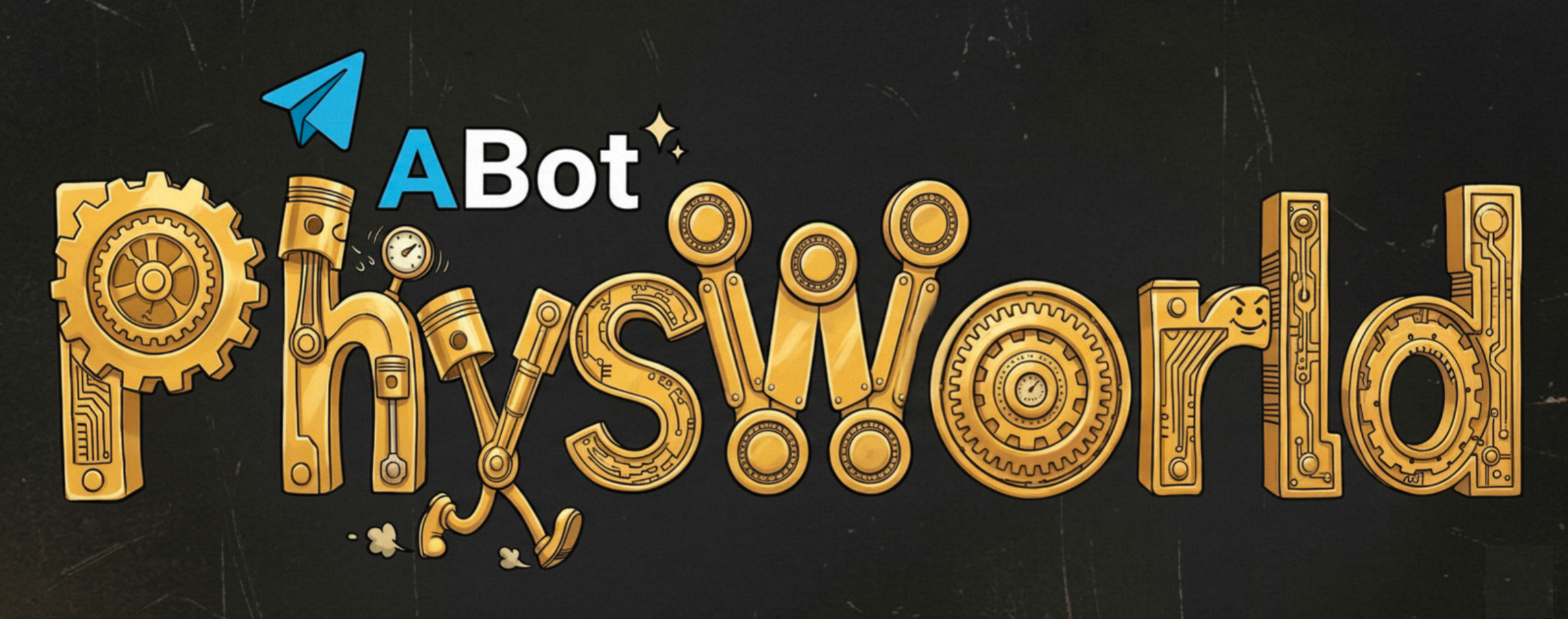}
    % \caption{\textbf{xxx}.}
    \label{fig:model}
\end{figure}

\vspace{-4pt}

\newpage
\tableofcontents
\newpage

\section{Introduction}
\label{sec:intro}

An embodied world model needs to generate future predictions that adhere to real-world physical laws in order to be effective for simulation, planning, and policy learning. Video generation presents a promising paradigm: such models can serve as simulators for Vision-Language-Action (VLA) policies~\cite{kim2024openvla,intelligence2025pi_,abot-m0, lingbot-vla}, provide interpretable trajectory previews, or function directly as World Action Models (WAMs)~\cite{ye2026worldactionmodelszeroshot, lingbot-va, kim2026cosmos-policy} by predicting action-conditioned dynamics—forming critical infrastructure for embodied intelligence.

Despite significant advances in visual fidelity, however, state-of-the-art models like Veo~3.1~\cite{google2026veo31} and Sora~v2~Pro~\cite{openai2025sora2} frequently produce manipulation sequences that violate basic physics, including object penetration, contactless motion, and unnatural deformations. These are not mere rendering artifacts but fundamental failures in physical reasoning, limiting their reliability in downstream robotic applications.

This gap arises from two core limitations: (i) training on general visual data lacking rich embodied interaction signals, which hinders the acquisition of fine-grained physical dynamics such as friction, collision response, and mass distribution; and (ii) reliance on standard maximum likelihood objectives during fine-tuning, which treat all prediction errors uniformly and fail to distinguish physically valid from invalid transitions. The absence of both embodied experience and physics-aware supervision results in a systematic disconnect between visual realism and physical plausibility.

To address this, we present \textbf{ABot-PhysWorld}, a physically grounded and action-controllable world model based on a 14B Diffusion Transformer~\cite{wan2025wan, wan2025open}, built upon a carefully designed data curation pipeline. We integrate three million real-world manipulation clips from five major open-source embodied datasets, enhancing data diversity and balance through curated sampling, ratio optimization, and physics-aware annotation—improving generalization across robots, objects, and environments. Building on this foundation, we introduce a physics-inspired DPO-based~\cite{qian2025rdpo,wang2025physcorr,cai2025phygdpo,qian2025rdpo} post-training framework with decoupled discriminators that suppress unphysical behaviors (e.g., object penetration, anti-gravity motion) while preserving visual quality and improving dynamic consistency. A parallel context block enables multi-channel spatial action injection, supporting precise cross-embodiment control and action-aligned motion synthesis. Together, ABot-PhysWorld generates visually realistic, physically plausible, and highly controllable manipulation sequences—serving as a high-fidelity interface for robot simulation and planning.

\begin{figure*}[htbp]
  \centering
  \includegraphics[width=\linewidth]{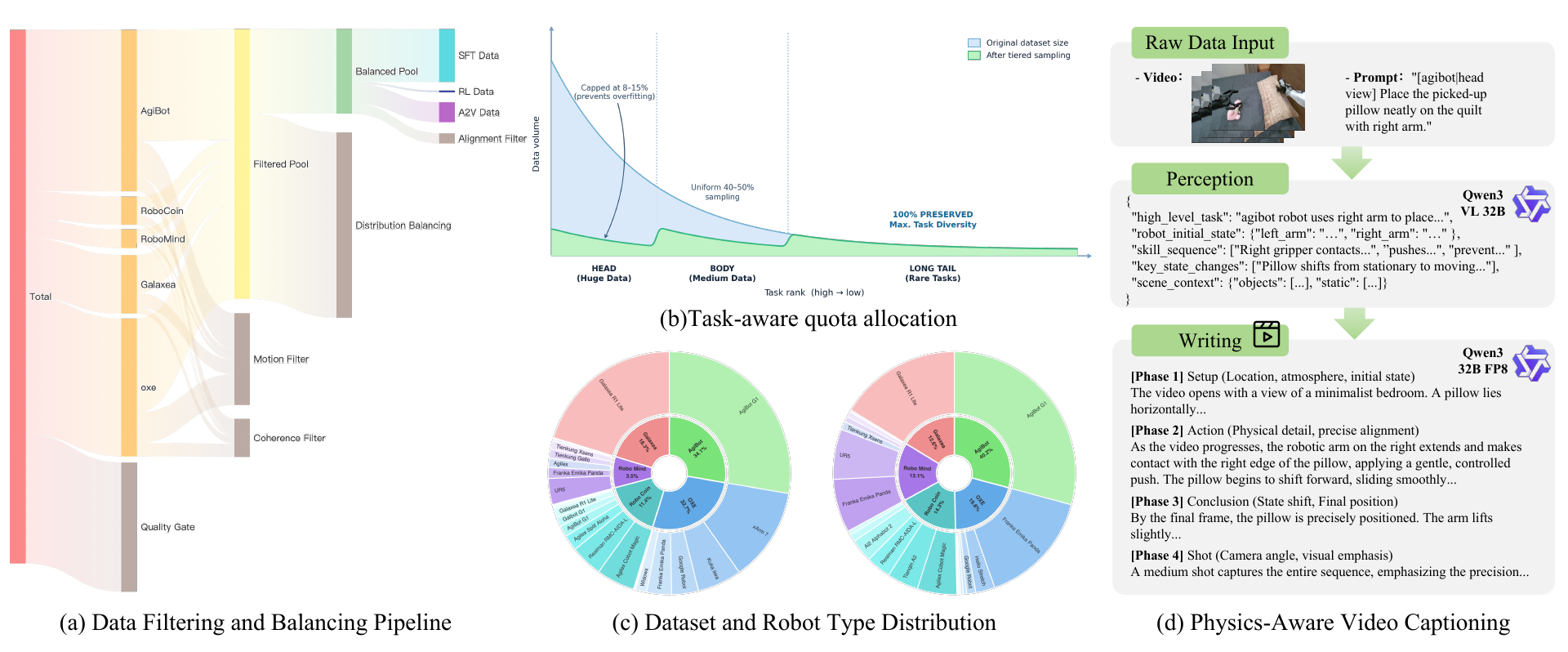}
  \vspace{-10pt}
  \caption{\textbf{Overview of the data curation pipeline.} (a) shows the multi-stage filtering and balancing flow from raw aggregation ($\sim$3M clips) to training-ready splits (SFT, RL, and A2V data). (b) Task-aware quota allocation: head tasks are capped at 8--15\%, body tasks are uniformly sampled at 40--50\%, and long-tail tasks are fully preserved to maximize task diversity. (c) Dataset and robot type distribution: the left ring shows the original composition and the right ring shows the rebalanced result after hierarchical sampling. (d) Physics-aware video captioning pipeline: a perception module (Qwen3-VL 32B) extracts structured physical attributes, followed by a writing module (Qwen3 32B FP8) that generates four-phase captions covering scene setup, action detail, state transition, and camera summary.}
  \label{fig:data_pipeline}
  \vspace{-5pt}
\end{figure*}

However, evaluating such advances remains challenging: existing benchmarks often emphasize visual quality or in-distribution accuracy, with little emphasis on physical consistency or zero-shot generalization. To enable more rigorous and realistic assessment, we propose \textbf{EZSbench}, the first training-independent \textbf{E}mbodied \textbf{Z}ero-\textbf{S}hot \textbf{Bench}mark combining both \textit{real} and \textit{synthetic} scenarios involving unseen combinations of robots, tasks, and scenes. Unlike existing benchmarks biased toward in-distribution fidelity, EZSbench is specifically designed to assess three key capabilities: action controllability, physical consistency, and zero-shot generalization. It employs a decoupled dual-model evaluation protocol that separately scores physical realism and action alignment, enabling fine-grained diagnosis of model behavior. We will publicly release EZSbench to promote standardized and meaningful progress in embodied video generation.

Our model achieves new state-of-the-art results on both PBench and EZSbench, surpassing Veo~3.1 and Sora~v2~Pro in physical plausibility and action trajectory consistency. More details are provided in Section~\ref{exp}. Our primary contributions are:

\begin{itemize}
    \item \textbf{Data}: We design a principled data curation pipeline that improves diversity and balance in embodied video data through curated sampling and physics-aware annotation—enabling scalable and robust training on real-world interactions.
    
    \item \textbf{Model}: We propose \textbf{ABot-PhysWorld}, a unified framework that jointly optimizes visual realism, physical plausibility, and action controllability through physics-aware DPO and parallel spatial action injection.
    
    \item \textbf{Evaluation}: We introduce \textbf{EZSbench}, the first training-independent zero-shot benchmark for embodied video generation, with a decoupled protocol to assess physical fidelity and action alignment under distribution shift.
\end{itemize}

% \section{Data Construction}
\section{Data Curation}

\begin{figure*}[t]
  \centering
  \includegraphics[width=\linewidth]{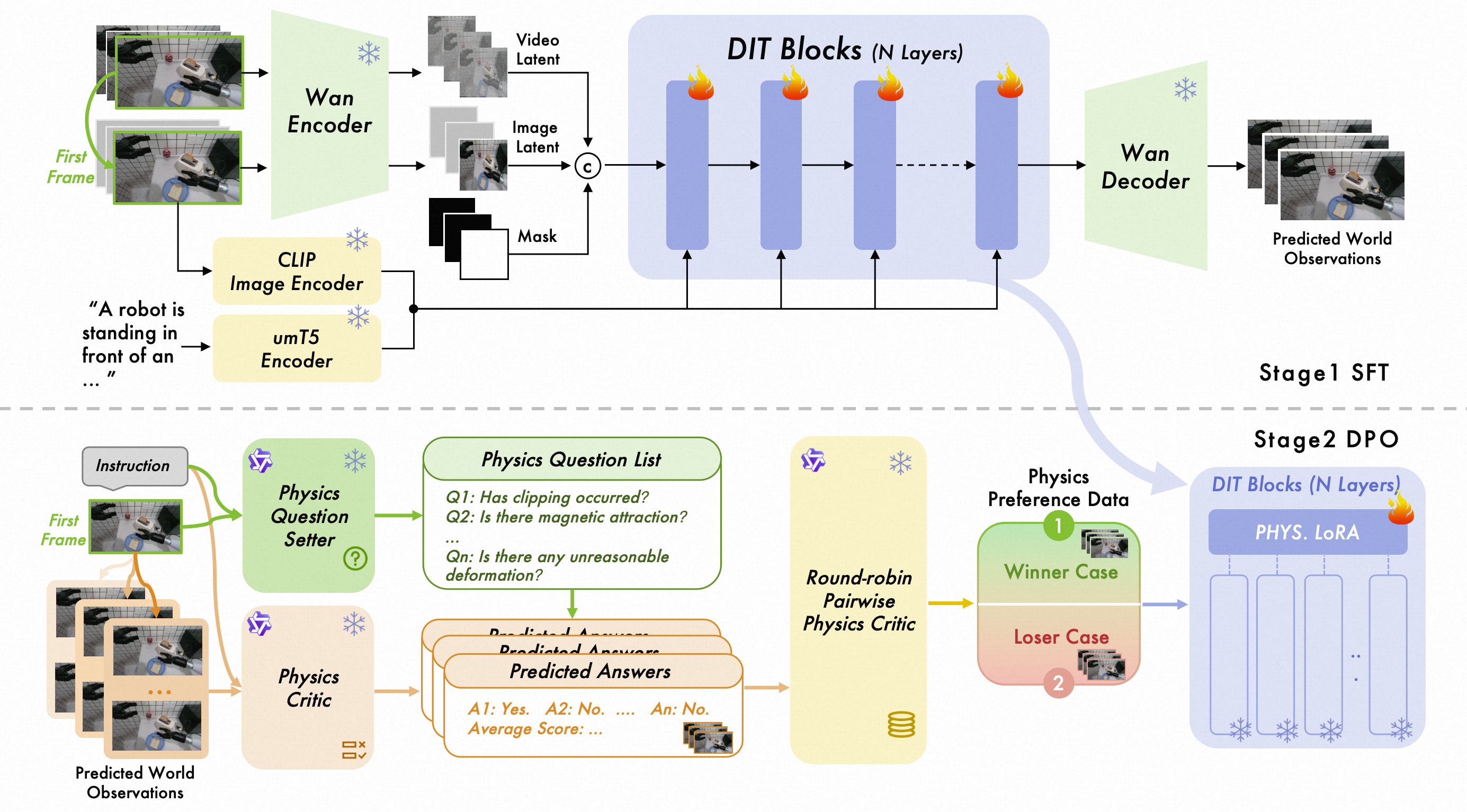}
    \vspace{-10pt}
  \caption{\textbf{Two-stage training pipeline.} \textbf{Stage~1}: SFT on the DiT to predict future frames from observations and instructions. \textbf{Stage~2}: generate $N$ candidates, score via physics checklist, and apply DPO via LoRA on frozen DiT weights.}
  \label{fig:Overview}
    \vspace{-5pt}
\end{figure*}

Data is essential for high-quality embodied world models.
% with scale, noise, and distributional representativeness jointly defining model knowledge and generalization. 
Following ~\cite{halevy2009unreasonable}, we adopt a data-driven approach through a systematic infrastructure that enhances data scale and diversity to address complex human-robot interaction modeling. As illustrated in Figure~\ref{fig:data_pipeline}, our data curation pipeline consists of three stages: embodied-specific filtering (\S\ref{sec:filtering}), hierarchical distribution balancing (\S\ref{sec:balancing}), and physically grounded caption generation (\S\ref{sec:captioning}).

\subsection{Embodied-Specific Data Filtering} 
\label{sec:filtering}
To build a physically consistent world model for embodied manipulation, we construct a foundational dataset of nearly three million real-world video clips by integrating five public datasets: 
% AgiBot~\cite{bu2025agibot} (1,000k), RoboCoin~\cite{wu2025robocoin} (336k), RoboMind~\cite{wu2024robomind} (104k), Galaxea~\cite{jiang2025galaxea} (537k), and OXE~\cite{o2024open} (960k).
AgiBot~\cite{bu2025agibot}, RoboCoin~\cite{wu2025robocoin}, RoboMind~\cite{wu2024robomind}, Galaxea~\cite{jiang2025galaxea}, and OXE~\cite{o2024open}.

General-domain curation pipelines such as Cosmos-Curate~\cite{cosmos2025curate} and VideoX-Fun~\cite{videox2024fun} are misaligned with embodied data: they rely on scene-cut detectors unsuitable for static-background manipulation videos, and prioritize visual aesthetics over physical causality. To resolve noise introduced by raw aggregation, we apply a video-level quality gate followed by three semantic filtering stages.

\noindent \textbf{Video-level quality gate.}  
Clips with abnormal resolutions or moving cameras are discarded. Sequences are constrained to 80--500 frames; longer videos are segmented temporally by task index into training-compliant clips to ensure relevance and efficiency.

\noindent \textbf{Optical-flow-based motion filtering.}  
We extract grayscale frames at 2 FPS and compute Farneb\"ack dense optical flow~\cite{farneback2003two} to capture pixel-level motion. By averaging the polar magnitudes of displacement vectors across each frame, we derive a global kinematic score and remove clips with near-zero motion or unphysical oscillations.

\noindent \textbf{CLIP-based temporal coherence.}  
To eliminate visual corruption (e.g., black screens, cuts, stitching errors), we assess temporal continuity using CLIP-based embeddings~\cite{radford2021learning}. Eight equidistant frames are sampled per clip, and their 768D features are extracted; samples with low average cosine similarity between consecutive frames are discarded.

\noindent \textbf{Vision-action alignment verification.}  
Calibrated action maps, encoding joint actions, end-effector poses, and gripper states, are projected onto video frames. Qwen3-VL verifies spatiotemporal alignment between visual motion and control signals, filtering out mismatches from sensor calibration or synchronization errors.

The resulting dataset provides a robust foundation for training generalizable, dynamics-aware world models across diverse embodied tasks.

\begin{figure}[!htbp]
  \centering
  \vspace{-10pt}
  \includegraphics[width=0.95\linewidth]{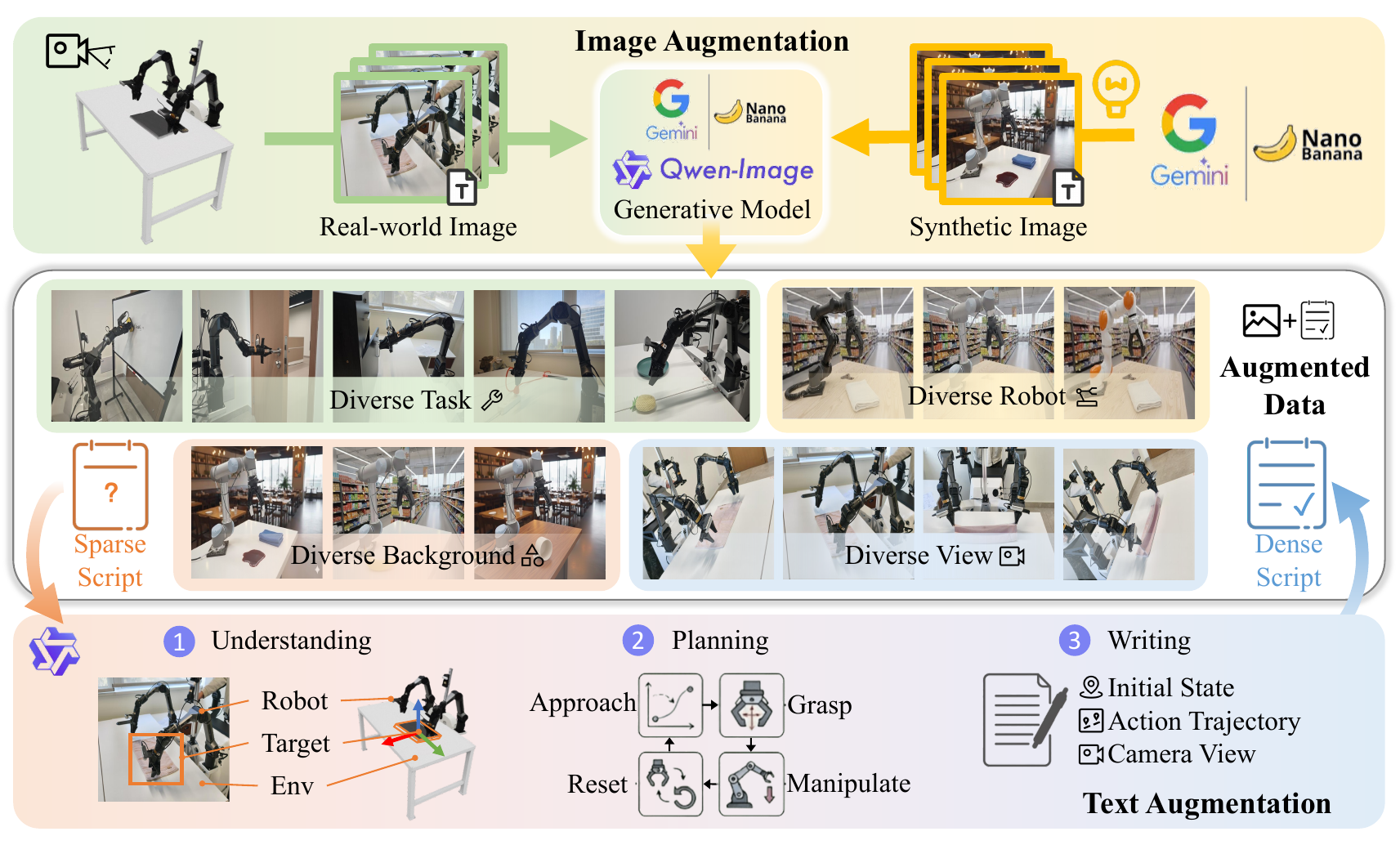}
  \vspace{-10pt}
  \caption{\textbf{Construction pipeline of the EZSbench.} \textbf{Top}: dual-source image augmentation---Branch~1 generates synthetic initial observations via text-to-image (Nano Banana) by varying robot morphology, scene, task, and viewpoint; Branch~2 applies VLM-guided background editing to real-world images while preserving foreground interactions. \textbf{Down}: three-stage dense description synthesis---visual anchoring grounds the scene layout and object coordinates, action simulation infers kinematically compliant trajectories with micro-physical interactions, and narrative synthesis produces a documentary-style caption integrating initial state, trajectory, and final state.}
  \vspace{-12pt}
  \label{fig:bench}
\end{figure}

\subsection{Hierarchical Distribution Balancing}
\label{sec:balancing}
Recent studies indicate that data diversity, not just volume, is key to scalable world models and generalist robotic policies~\cite{kang2024how,ye2026worldactionmodelszeroshot}; scaling repetitive data often leads to memorization rather than out-of-distribution generalization. To address this, we design a hierarchical dynamic sampling strategy spanning four levels: video, sub-dataset or robot type, task, and macro-dataset. This approach balances data distribution while preserving long-tail features.

\noindent \textbf{Level 1: Intra-dataset diversity preservation.}  
Several source datasets are themselves aggregations of smaller collections; for example, small sub-datasets within OXE~\cite{o2024open} are retained entirely to preserve unique interaction patterns before any cross-dataset operations are applied.

\noindent \textbf{Level 2: Cross-robot rebalancing.}  
Operating across the five source datasets, this level addresses imbalances among robot embodiment types. Underrepresented robot types are upweighted to retain rare interaction patterns (e.g., non-standard kinematics or dual-arm coordination), enhancing cross-platform generalization and mitigating head-category dominance (Figure~\ref{fig:data_pipeline}c).

\noindent \textbf{Level 3: Task-aware quota allocation.}  
Rather than applying a fixed sampling threshold, we partition tasks into three tiers based on data volume and assign tier-specific strategies (Figure~\ref{fig:data_pipeline}b). Head tasks (high data volume) are capped at 8--15\% of their original size to prevent overfitting to dominant categories. Body tasks (medium volume) are uniformly sampled at 40--50\%, preserving representative coverage without excessive redundancy. Long-tail tasks (rare tasks) are fully preserved to maximize task diversity. 

\noindent \textbf{Level 4: Macro-dataset scale regulation.}  
Finally, at the coarsest granularity, large-scale macro-datasets (e.g., AgiBot, OXE) are capped via uniform subsampling 
% to prevent any single source from dominating the training distribution
, while micro-datasets (e.g., RoboMind) are guaranteed minimum coverage through a mandatory lower bound. When activated, a three-round supplementation strategy allocates: (1) a base quota uniformly across tasks, (2) reallocates unused quotas proportionally, and (3) fills residual gaps via random fallback sampling from the global filtered pool, preventing single-task over-extraction and improving long-tail balance.
This hierarchical framework improves combinatorial diversity and distributional balance, providing a robust foundation for training general-purpose embodied world models.

\subsection{Physics-Aware Video Captioning} 
\label{sec:captioning}
Training embodied world models with text-to-video (T2V) objectives requires captions that go beyond surface-level scene descriptions. An effective annotation must capture three progressively deeper aspects of robotic manipulation: \emph{what} the robot does (action semantics), \emph{how} it interacts with the physical world (spatial and contact precision), and \emph{why} the observed outcome occurs (causal reasoning). We design a multi-level annotation system that addresses each of these requirements.

\noindent \textbf{Multi-level action semantics.}  
Adopting a robot-centric ``annotating for action'' philosophy, we structure each caption across four granularities: macroscopic task intent in natural language; mesoscopic verb-noun action segmentation for long-horizon planning; microscopic details including Cartesian trajectories, relative motion, and gripper states; and scene-level descriptions of physical relations (contact, support, containment) and task outcomes (success, failure, partial accidents).

\noindent \textbf{Grounded spatial precision.}  
Purely template-driven annotation tends to produce hallucinated spatial relations and imprecise grasp descriptions. To suppress these errors, we introduce three mechanisms: few-shot in-context learning with explicit positive and negative examples for richer physical detail, dynamic vocabularies for precise grasp-type specification, and a visible-fact baseline that restricts descriptions to observable evidence.

\noindent \textbf{Causal physical modeling.}  
Beyond describing what happens, effective captions for world models must explain why it happens. We explicitly annotate physical causality, including gravity-induced dropping, surface deformation, and force feedback. A four-stage narrative structure (scene construction, action flow, final state confirmation, and camera summary) organizes each caption into a temporally coherent account of the manipulation episode.

This annotation system delivers physically grounded language supervision that captures not only events but their underlying causes, providing the semantic foundation for training world models with causal understanding.

\section{Method}

\begin{table*}[t]
\centering
\vspace{-10pt}
\caption{Quantitative comparison on the PAI-Bench robot domain subset.}
\label{tab:quantitative_results}
  \vspace{-8pt}

\resizebox{\textwidth}{!}{
\begin{tabular}{lccccccccccc}
\toprule
Model & AQ & BC & IQ & MS & OC & SC & I2VB & I2VS & Quality Score & Domain Score & Avg. \\
\midrule
Wan 2.5 & \textbf{0.5477} & 0.8985 & 0.6458 & 0.9623 & 0.2190 & 0.8784 & 0.9438 & 0.9428 & 0.7548 & 0.8644 & 0.8096 \\
GigaWorld-0 & 0.4757 & 0.9219 & 0.6506 & 0.9908 & 0.1944 & 0.9111 & 0.9673 & 0.9607 & 0.7591 & 0.8583 & 0.8087 \\
Veo 3.1 & 0.5458 & 0.9216 & \textbf{0.7244} & 0.9712 & \textbf{0.2214} & 0.9146 & 0.9317 & \textbf{0.9614} & \textbf{0.7740} & 0.8350 & 0.8045 \\
Wan2.1\_14B & 0.4723 & 0.9315 & 0.7118 & \textbf{0.9917} & 0.1921 & 0.9185 & 0.9745 & 0.9451 & 0.7672 & 0.8391 & 0.8032 \\
WoW-wan 14B & 0.4664 & 0.9295 & 0.7027 & 0.9858 & 0.1941 & 0.9149 & 0.9613 & 0.9292 & 0.7605 & 0.8301 & 0.7953 \\
Cosmos-Predict 2.5 & 0.4897 & 0.9166 & 0.7405 & 0.9906 & 0.1911 & 0.8973 & 0.9304 & 0.9030 & 0.7574 & 0.8021 & 0.7797 \\
Sora v2 Pro& 0.5324 & 0.9285 & 0.6956 & 0.9702 & 0.2203 & 0.9163 & 0.9507 & 0.9290 & 0.7679 & 0.7626 & 0.7652 \\
UnifoLM-WMA-0 & 0.4547 & \textbf{0.9423} & 0.6564 & 0.9875 & 0.1878 & \textbf{0.9412} & 0.9638 & 0.9403 & 0.7593 & 0.6693 & 0.7143 \\
Our Model & 0.4620 & 0.9373 & 0.6906 & 0.9916 & 0.1927 & 0.9406 & \textbf{0.9777} & 0.9498 & 0.7678 & 0.8785 & 0.8232 \\
Our Model + DPO & 0.4667 & 0.9365 & 0.6916 & 0.9908 & 0.1942 & 0.9355 & 0.9768 & 0.9483 & 0.7676 & \textbf{0.9306} & \textbf{0.8491} \\
\bottomrule
\end{tabular}
}
\vspace{-8pt}
\end{table*}

\subsection{Embodied Video Generation Backbone}
Generating physically plausible manipulation videos requires a backbone that captures both the visual diversity of real-world scenes and the fine-grained spatiotemporal dynamics of robot-object interactions. To meet this requirement, we build upon Wan2.1-I2V-14B~\cite{wan2025open}, 
% a 14-billion-parameter image-to-video Diffusion Transformer (DiT) pre-trained on large-scale natural video data, 
and fully fine-tune it on our curated embodied dataset. 
% We crop and resize inputs to $480\times832$, sample 81 frames uniformly, and encode them with a pre-trained VAE.

% To condition generation on the initial observation, we concatenate the first-frame VAE encoding with a binary mask. Cross-attention integrates T5-XXL~\cite{raffel2020exploring} text instruction encodings, and 3D Rotary Position Embedding~\cite{su2024roformer} (RoPE) encodes the spatiotemporal coordinates.

% The model is optimized with flow matching~\cite{lipman2022flow} and decoded via Classifier-Free Guidance~\cite{ho2022classifier} at inference.

\subsection{Physical Preference Alignment}

While SFT teaches the model to reproduce training distributions, it treats all samples equivalently and cannot distinguish physically correct predictions from those containing violations such as object penetration or anti-gravity motion. To explicitly suppress these violations, we propose a post-training preference alignment pipeline (Figure~\ref{fig:Overview}) that pairs a decoupled VLM discriminator with Diffusion-DPO.

\subsubsection{Decoupled VLM Discriminator}
For a given prompt $x$ and initial state, we generate $N$ candidate video variants. Evaluating physical plausibility with a single VLM risks self-evaluation hallucinations, where the same model that generates questions also judges answers. To prevent this, we decouple the evaluation into two roles.

The Qwen3-VL 32B Thinking model acts as the \emph{proposer}. It observes the first frame and text instruction to dynamically generate a task-specific physical checklist based on a hierarchical evaluation system. This system applies single-vote veto power to Tier~1 metrics (fatal violations such as penetration and anti-gravity) and uses Tier~2 metrics (micro-physical fidelity and contact dynamics) to differentiate compliant samples. Generating specific questions prevents hallucinations caused by vague queries. For example, given the instruction to grasp and place an apple, the proposer asks whether the gripper penetrates the apple, whether the apple penetrates the bag, and whether it is firmly grasped rather than magnetically attached. The proposer also explicitly constructs a balanced mix of positive and negative questions to prevent the scoring model from sycophantically predicting the absence of violations.

The Gemini 3 Pro model~\cite{gemini3_2025} then acts as the \emph{scorer}. It uses explicit Chain-of-Thought reasoning, including global scanning, marking suspicious frames, and backtracking confirmation, to evaluate the $N$ variants against the generated checklist. To efficiently resolve score ties and isolate the optimal ($y_w$) and worst ($y_l$) samples within $\mathcal{O}(N)$ complexity, we apply a multi-round tournament-based sampling strategy: a knockout tournament first selects the optimal sample, followed by a loser-bracket round to identify the worst sample. This two-stage mechanism avoids full permutation comparisons and yields highly discriminative DPO~\cite{rafailov2023direct} training triplets $(x, y_w, y_l)$ with clear margins.

\subsubsection{Diffusion-DPO Training}
Given the discriminative triplets $(c, v_w, v_l)$ produced by the decoupled discriminator, where $c$ is the condition, $v_w$ the physics-compliant video, and $v_l$ the physics-violating video, we adapt the Diffusion-DPO framework to fine-tune the video diffusion model directly in the latent space. For a video latent $z$, we inject Gaussian noise $\epsilon \sim \mathcal{N}(0, I)$ at time step $t \sim \mathcal{U}(0, T)$ to obtain $z_t$. The single-step denoising mean squared error for model $\epsilon_\theta$ is $L(\theta, z) = \| \epsilon_\theta(z_t, t, c) - \epsilon \|_2^2$. Letting $L_\theta(\cdot)$ and $L_{ref}(\cdot)$ denote the denoising errors of the policy model $\pi_\theta$ and reference model $\pi_{ref}$ (the SFT baseline) respectively, the physical preference alignment loss is:
\begin{multline}
    \mathcal{L}_{DPO} = - \mathbb{E}_{z, \epsilon, t} \Bigg[ \log \sigma \Bigg( 
    -\frac{\beta}{2} \Big[ 
    \underbrace{(L_\theta(z_w) - L_\theta(z_l))}_{\text{Policy Diff.}} \\
    - \underbrace{(L_{ref}(z_w) - L_{ref}(z_l))}_{\text{Ref.\ Diff.}} 
    \Big] \Bigg) \Bigg],
\end{multline}
where $\beta$ controls distribution divergence, and $z_w, z_l$ are the latents of $v_w, v_l$. This objective actively reduces the prediction error for $z_w$ while increasing it for $z_l$ at each timestep.

Standard DPO requires maintaining two complete computation graphs ($\pi_\theta$ and $\pi_{ref}$), causing out-of-memory errors for a 14B DiT. To resolve this, we freeze the DiT backbone and inject Low-Rank Adaptation~\cite{hu2021lora} (LoRA) modules with a rank of 64 into the self-attention (query, key, value, output) and feed-forward layers, so that the reference model loss $L_{ref}$ can be computed by temporarily disabling the LoRA weights with zero additional memory.

\subsection{Action-Conditioned Video Generation}
Beyond text-conditioned prediction, a world model for embodied intelligence must support controllable generation: given the current observation and a future action sequence, it should produce physically plausible videos that faithfully follow the commanded trajectory. Directly injecting low-dimensional robotic commands (\eg, end-effector poses) into high-dimensional visual pipelines creates a semantic gap. To bridge this gap, we convert discrete action commands into spatially structured action maps and inject them through parallel context blocks that preserve the backbone's pre-trained physical knowledge.

\subsubsection{Action Map Construction}
The input action is a 7D vector $\boldsymbol{a}\in \mathbb{R}^7$ (3D position, 3D orientation, gripper openness), extending to 14 dimensions for dual-arm systems. Using camera intrinsics and extrinsics, we project the 3D position $(x,y,z)$ to a 2D center $(u, v)$. Orientation is encoded as the three principal axes of the corresponding rotation matrix, projected into the image plane and rendered as colored arrows whose length encodes depth. The gripper state is mapped to a circular mask at $(u, v)$, with opacity linearly indicating openness. For dual-arm robots, we distinguish left and right arms via red and blue channels, yielding a multi-channel action map.

\subsubsection{Action Injection}
Existing action injection methods either use Adaptive Layer Normalization (AdaLN)~\cite{dit} for MLP-encoded actions~\cite{cosmos,zhu2025irasim}, which hinders cross-embodiment generalization, or concatenate action maps directly with noisy latents for full fine-tuning~\cite{evac, liao2025genie, team2025gigaworld}, causing catastrophic forgetting of pre-trained physical priors.

\begin{figure}[htbp]
  \centering
  \vspace{-8pt}
  \includegraphics[width=0.99\linewidth]{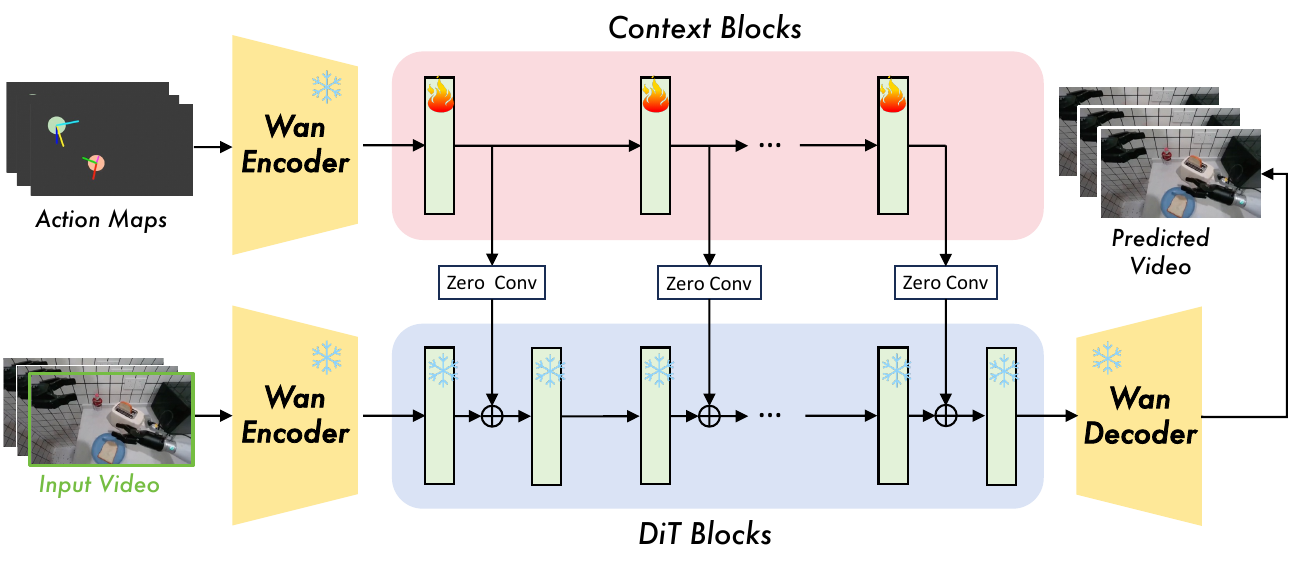}
  \vspace{-10pt}
  \caption{Architecture of the action-conditioned video generation model. We selectively duplicate DiT blocks as parallel context blocks to process action maps, and fuse their outputs residually into the main DiT.
}
  \label{fig:a2v_model}
  \vspace{-10pt}
\end{figure}

To address these challenges, as shown in Figure~\ref{fig:a2v_model}, we clone selective blocks from the main DiT~\cite{wan2025wan} to form a parallel set of context blocks that process the action maps~\cite{vace}. The output of each context block is projected via zero-initialized convolution layers and added residually to the corresponding main DiT block:
\begin{equation}
    \mathbf { x } _ { i } = { \mathrm {DiT}}_ {i} (\mathbf{x} _{i-1}) + \alpha\cdot W_{\mathrm {zero}} ^ {( i )} \mathbf{h}_{i},
\end{equation}
where $\mathbf{h}_i$ is the $i$-th context block output, $W_{\mathrm{zero}}$ is the zero-initialized convolution layer, and $\alpha$ is the control scale. Following VACE, we instantiate context blocks selectively, replicating only every fifth DiT block. Because the zero initialization ensures that the context branch contributes no signal at the start of training, the backbone weights remain undisturbed, preserving pre-trained physical priors while gradually learning action controllability.

\begin{table*}[t]
\centering
\caption{Quantitative comparison on EZSbench.}
\label{tab:real_data_v6}
  \vspace{-8pt}
\resizebox{\textwidth}{!}{
\begin{tabular}{lccccccccccc}
\toprule
Model & AQ & BC & IQ & MS & OC & SC & I2VB & I2VS & Quality Score & Domain Score & Avg. \\
\midrule
WoW-wan 14B & 0.4764 & 0.9412 & \textbf{0.7514} & 0.9922 & 0.1236 & 0.9347 & 0.9495 & 0.9178 & 0.7609 & 0.7951 & 0.7780 \\
GigaWorld-0 & 0.4466 & 0.8893 & 0.6565 & 0.9889 & 0.1222 & 0.8678 & 0.9325 & 0.9139 & 0.7272 & 0.7826 & 0.7549 \\
Cosmos-Predict 2.5 & 0.4148 & 0.8835 & 0.6810 & 0.9878 & 0.0970 & 0.8366 & 0.9054 & 0.8653 & 0.7089 & 0.7698 & 0.7394 \\
UnifoLM-WMA-0 & \textbf{0.4861} & \textbf{0.9575} & 0.7390 & \textbf{0.9925} & 0.1046 & \textbf{0.9452} & 0.8421 & 0.8170 & 0.7355 & 0.5232 & 0.6294 \\
Our Model & 0.4789 & 0.9494 & 0.7483 & 0.9910 & \textbf{0.1301} & 0.9442 & \textbf{0.9680} & \textbf{0.9453} & \textbf{0.7694} & \textbf{0.8366} & \textbf{0.8030} \\
\bottomrule
\end{tabular}
}
\end{table*}

\begin{figure*}[t]
  \centering
  \vspace{-3pt}
  \includegraphics[width=0.9\linewidth]{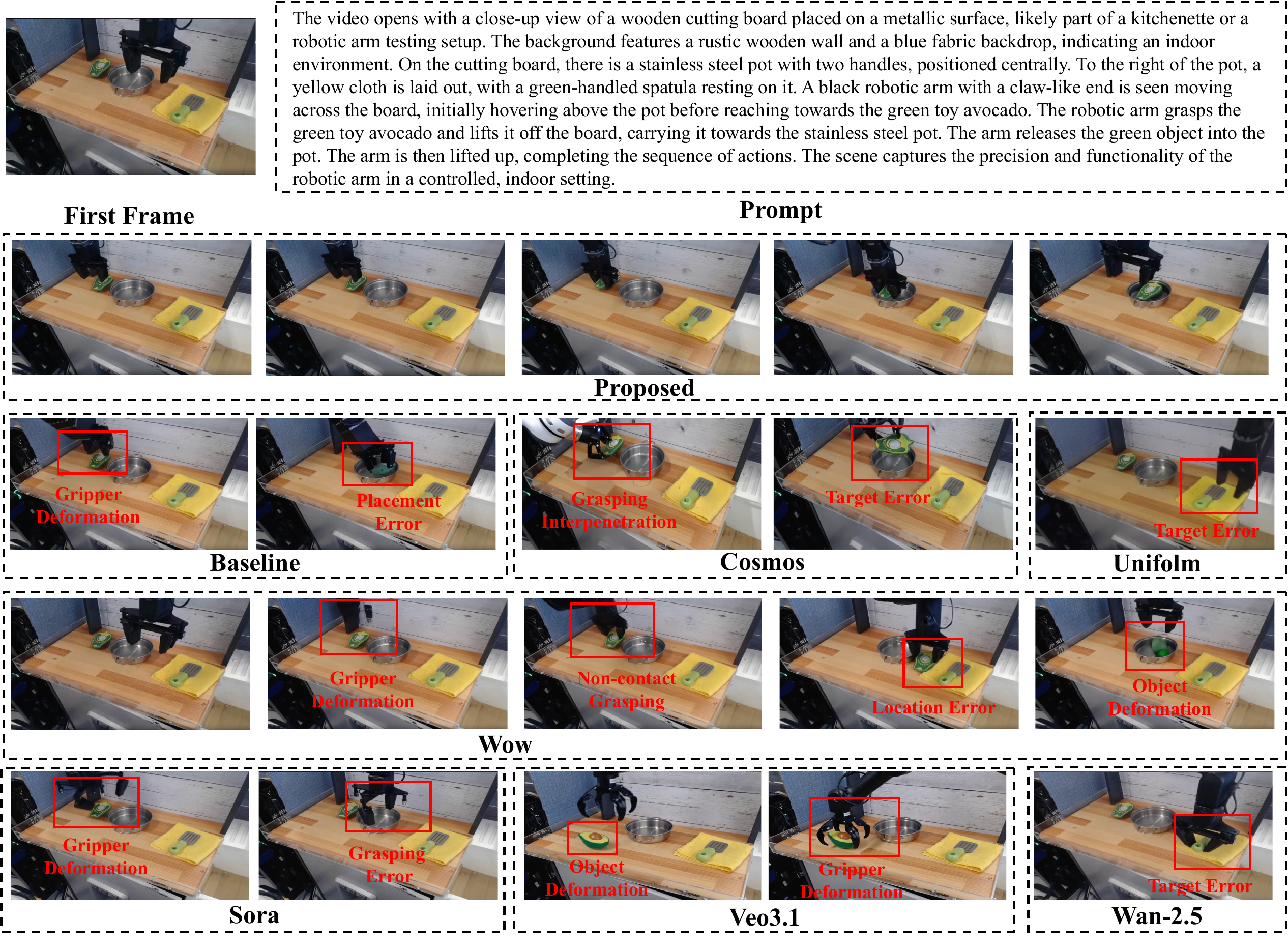}
  \vspace{-8pt}
  \caption{Qualitative comparison on PAI-Bench.}
  \label{fig:image-2.png}
  \vspace{-5pt}
\end{figure*}
\section{Embodied-ZeroShot Benchmark}
\label{sec:benchmark}

Existing embodied video generation benchmarks draw test samples from the same distribution as training data, making it difficult to assess genuine zero-shot generalization. We introduce EZSbench to evaluate physical fidelity and cross-embodiment generalization under fully out-of-distribution conditions, where diverse robot morphologies, environments, and tasks are composed into previously unseen combinations with no overlap with training data.

\subsection{Evaluation Set}
We construct the initial observation pool via a dual-branch strategy. The first branch generates synthetic images using the text-to-image model Nano Banana~\cite{google2025nanobanana}, controlled by four orthogonal variables: robots, scenes, tasks, and perspectives. This approach targets morphological, scene, and task generalization by varying arm structures, backgrounds, and task complexity from basic pick-and-place to long-horizon manipulations. The second branch uses a large VLM for controllable scene editing on real-world mechanical arm images, dynamically altering backgrounds while preserving foreground physical interactions.

To generate reliable physical descriptions, we propose a physics-heuristic dense description synthesis framework that progresses through visual anchoring, kinematically compliant action simulation, and narrative synthesis, producing text that integrates the initial state, action trajectory, and final state. Each initial image paired with its dense description forms a core benchmark sample.

\subsection{Evaluation Method}
A key challenge in evaluating physical consistency is that a single model acting as both question generator and answer judge introduces self-evaluation bias. To address this, we propose a decoupled dual-model evaluation paradigm. The Qwen3-VL-32B-Thinking model dynamically generates physical checklist questions based on the initial state and text instructions. It employs a System~2 reasoning protocol for scene decoding and action parsing, guided by few-shot examples covering nine criteria across spatial, temporal, and physical dimensions. We mandate that 30--50\% of the checklist comprises negative questions, such as asking whether a red apple is green, to prevent shortcut learning via random guessing. To eliminate self-evaluation bias, the Qwen2.5-VL-72B-Instruct model serves as the answering end. The final physical score $S_v$ for a video $v$ measures the consistency between the visual question answering (VQA) predictions and the checklist ground truth (GT):
$S_v = \frac{1}{|Q_v|} \sum_{q \in Q_v} \mathbb{I}(\text{VQA}(v, q) = \text{GT}(q))$, where $Q_v$ is the generated checklist and $\mathbb{I}(\cdot)$ is the indicator function.
\section{Experiments}
\label{exp}

\subsection{Implementation Details}

We conduct all experiments on a cluster of 128 Nvidia H20 GPUs. The training pipeline consists of three stages: TI2V foundational training, DPO, and A2V training.
\textbf{For TI2V}, we use Wan2.1-I2V-14B-480P with inputs cropped to $480\times832$ and 81 frames uniformly sampled. Trained for 6,000 steps with a global batch size of 128 and learning rate 1e-5.
\textbf{For DPO}, we apply LoRA-based fine-tuning to the Diffusion Transformer, inserting rank-64 adapters (scaling factor 64) into self-attention and feed-forward layers (q, k, v, o, ffn.0, ffn.2). Optimized with AdamW, lr=1e-6, 10-step warmup. To strengthen preference signals in diffusion models, we set $\beta = 5000$. Training employs BF16 mixed precision, gradient checkpointing, per-device batch size 1, and runs for 500 steps/epoch over 100 epochs.
\textbf{For A2V}, we adopt the VACE framework on the fine-tuned TI2V model. We duplicate specific Diffusion Transformer layers (0, 5, 10, 15, 20, 25, 30, 35) as a trainable context branch while keeping the backbone frozen. Data is augmented via random frame sampling with variable stride. Training uses batch size 16, learning rate 5e-5, and 20,000 steps.

% We deploy all experiments on a computing cluster of 128 Nvidia H20 GPUs. The training pipeline comprises three stages: TI2V foundational training, DPO, and A2V training.

% For TI2V training, we use Wan2.1-I2V-14B-480P. Inputs are cropped to $480\times832$, retaining 81 frames via uniform sampling. Training runs for 6000 steps with a global batch size of 128 and a learning rate of 1e-5.

% For DPO, we apply parameter-efficient fine-tuning via LoRA on the Diffusion Transformer. We inject rank-64 adapters (scaling factor 64) into the self-attention and specific feed-forward layers (q, k, v, o, ffn.0, ffn.2). Optimization uses AdamW (learning rate 1e-6) with a 10-step warmup. To obtain a sufficient preference gradient for diffusion models, we set the preference strength $\beta = 5000$. Training uses BF16 mixed precision, gradient checkpointing, a per-device batch size of 1, and runs for 500 steps per epoch across 100 epochs.

% For A2V training, we apply the VACE framework to the fine-tuned TI2V base model. We duplicate specific Diffusion Transformer layers (0, 5, 10, 15, 20, 25, 30, 35) as a context block branch, which undergoes fine-tuning while the backbone remains frozen. A random frame sampling strategy with random stride augments the data. Training proceeds with a batch size of 16 and a learning rate of 5e-5 for 20000 steps.

\subsection{Evaluation Setup}

% \subsubsection{Text-Conditioned Generation}
\noindent \textbf{Text-Conditioned Generation.}  
We evaluate physical plausibility and visual quality using PAI-Bench~\cite{zhou2025paibench} and its PBench dataset, focusing on the robot domain subset with 174 complex manipulation videos from BridgeData~V2~\cite{walke2023bridgedata}, AgiBot, and Open X-Embodiment. We employ an MLLM-as-Judge approach with Qwen2.5-VL-72B-Instruct~\cite{bai2025qwen25vl} for binary visual question answering. The Domain Score evaluates accuracy across 886 questions in three dimensions: spatial (36.3\%, geometry and contact), temporal (28.6\%, causal logic), and physical (34.1\%, object attributes and state changes). We also use PAI-Bench’s multidimensional quality metrics: subject~\cite{caron2021emerging}/background ~\cite{fu2023dreamsim} consistency, Overall consistency \cite{wang2024internvid}, Aesthetic quality (LAION aesthetic head), Imaging quality ~\cite{ke2021musiq}, motion smoothness, and i2v subject/background consistency. For zero-shot evaluation, we use EZSbench with the decoupled dual-model protocol from Section~\ref{exp}.

% We evaluate physical plausibility and visual quality using the PAI-Bench~\cite{zhou2025paibench} framework and PBench dataset. We focus on the PAI-Bench robot domain subset, containing 174 complex manipulation videos from BridgeData~V2~\cite{walke2023bridgedata}, AgiBot, and Open X-Embodiment. We adopt an MLLM-as-Judge paradigm with Qwen2.5-VL-72B-Instruct~\cite{bai2025qwen25vl} for binary visual question answering. The Domain Score measures accuracy across 886 questions in three dimensions: spatial (36.3\%, geometry and contact), temporal (28.6\%, causal logic), and physical (34.1\%, object attributes and state changes). We also apply PAI-Bench's multidimensional quality metrics: Subject (DINO ViT-S/16~\cite{caron2021emerging}) for foreground semantic integrity, Background (DreamSim~\cite{fu2023dreamsim}) for layout stability, Overall consistency (ViCLIP~\cite{wang2024internvid}) for text-video alignment, Aesthetic quality (LAION aesthetic linear head + CLIP), and Imaging quality (MUSIQ~\cite{ke2021musiq}), along with motion smoothness (MS) and subject/background consistency (SC/BC). For zero-shot evaluation, we use our proposed EZSbench with the decoupled dual-model evaluation protocol described in Section~5.

% \subsubsection{Action-Conditioned Generation}
\noindent \textbf{Action-Conditioned Generation.}
We construct the action-conditioned evaluation set by uniformly sampling 200 instances from the action-to-video dataset, each containing an initial frame and a structured action sequence (end-effector pose and gripper state). Visual alignment is evaluated frame-by-frame using PSNR for pixel accuracy and SSIM for local texture fidelity. For trajectory accuracy, we use nDTW: a fine-tuned YOLO detector locates the gripper in each frame, and the extracted trajectory is compared to ground truth via nDTW.
% We construct the action-conditioned evaluation set by uniformly sampling 200 instances from the action-to-video dataset. Each sample contains an initial frame and a structured action sequence specifying end-effector pose and gripper state. We evaluate visual alignment frame-by-frame using Peak Signal-to-Noise Ratio (PSNR) for pixel accuracy and Structural Similarity (SSIM) for local texture fidelity. For trajectory accuracy, we use normalized dynamic time warping (nDTW): a fine-tuned YOLO detector identifies the robot gripper in each frame, and the extracted trajectory is compared against the ground-truth video using nDTW.

% \subsubsection{Baselines}
\noindent \textbf{Baselines.}
We compare with Cosmos-Predict~2.5-2B~\cite{agarwal2025cosmos}, GigaWorld-0~\cite{gigaai2025gigaworld}, UnifoLM-WMA-0~\cite{unitree2025unifolm}, WoW-wan~14B~\cite{chi2025wow}, Veo~3.1~\cite{google2026veo31}, Sora~v2~Pro~\cite{openai2025sora2}, and Wan~2.5~\cite{alibaba2025wan25} for text-conditioned generation, and with Enerverse-AC~\cite{evac} and Gen-Sim~\cite{liao2025genie} for action-conditioned generation.

% We compare against Cosmos-Predict~2.5-2B~\cite{agarwal2025cosmos}, GigaWorld-0~\cite{gigaai2025gigaworld}, UnifoLM-WMA-0~\cite{unitree2025unifolm}, WoW-wan~14B~\cite{chi2025wow}, Veo~3.1~\cite{google2026veo31}, Sora~v2~Pro~\cite{openai2025sora2}, and Wan~2.5~\cite{alibaba2025wan25} for text-conditioned generation, and against Enerverse-AC and Gen-Sim for action-conditioned generation.

\begin{table}[!htbp]
\centering
\caption{Quantitative results on action-conditioned generation.}
% \vspace{-8pt}
\label{tab:quantitative_results_a2v}
\begin{tabular}{lccc}
\toprule
Model & PSNR & SSIM & Traj. Consis.\\
\midrule
Enerverse-AC & 20.42 & 0.7542 & 0.8157 \\
Gen-Sim & 18.05 & 0.7413 & 0.6195 \\
Ours & \textbf{21.09} & \textbf{0.8126} & \textbf{0.8522}\\
\bottomrule
\end{tabular}
\end{table}

% \subsection{Text-Conditioned Generation Results}
\subsection{Evaluation Results}

% \subsubsection{PBench Evaluation}
\noindent \textbf{PBench Evaluation.}
As shown in Table~\ref{tab:quantitative_results}, our DPO-augmented model achieves the highest average score (0.8491) and sets a new state-of-the-art Domain Score (0.9306), outperforming the base model (0.8785) and all baselines. Existing methods show a trade-off between visual quality and physical fidelity: Veo~3.1 and Sora~v2~Pro achieve high Quality Scores (0.7740, 0.7679) due to strong imaging and aesthetics, but lag in Domain Score (0.8350, 0.7626), favoring perception over physics. Our model maintains competitive visual quality (Quality Score: 0.7676) while enforcing physical constraints, proving that alignment with physical laws does not sacrifice perceptual quality. The base model also shows strong spatiotemporal stability (I2VB: 0.9777; MS: 0.9916).
% As shown in Table~\ref{tab:quantitative_results}, our DPO-augmented model achieves the highest average score (0.8491) and establishes a state-of-the-art Domain Score (0.9306), outperforming both our base model (0.8785) and all baselines. Existing methods exhibit a clear trade-off between visual quality and physical fidelity: Veo~3.1 and Sora~v2~Pro achieve competitive Quality Scores (0.7740 and 0.7679) due to high imaging and aesthetic metrics, but score lower on Domain Scores (0.8350 and 0.7626), prioritizing visual perception over physical laws. Our model maintains competitive visual quality (Quality Score: 0.7676) while strictly enforcing physical constraints, demonstrating that physical alignment need not come at the expense of perceptual quality. The base model also exhibits high spatiotemporal stability (I2VB: 0.9777; MS: 0.9916).

% \subsubsection{EZSbench Evaluation}
\noindent \textbf{EZSbench Evaluation.}
On the out-of-distribution EZSbench (Table~\ref{tab:real_data_v6}), our model achieves the highest overall average score (0.8030), establishing state-of-the-art results for both Quality Score (0.7694) and Domain Score (0.8366). This confirms that the physical fidelity improvements generalize beyond the training distribution.

% \subsubsection{Qualitative Analysis}
\noindent \textbf{Qualitative Analysis.}
Figure~\ref{fig:image-2.png} shows qualitative PBench comparisons. Baselines violate physical laws in complex interactions: Sora~v2~Pro and Veo~3.1 show gripper or object distortion during dense contact; GigaWorld-0 and Cosmos exhibit grasping penetration; WoW produces non-contact grasping and geometric distortion; UnifoLM and Wan~2.5 misidentify targets (e.g., spatula instead of rag). Our method correctly identifies targets, maintains spatiotemporal coherence, and avoids deformation and penetration.
% 
% Figure~\ref{fig:image-2.png} presents qualitative comparisons on PBench. Baselines frequently violate physical laws during complex interactions: Sora~v2~Pro and Veo~3.1 display gripper or object distortion during high-density contact, GigaWorld-0 and Cosmos exhibit grasping penetration, WoW shows non-contact grasping and geometric distortion, and UnifoLM and Wan~2.5 misidentify the target (\eg, a spatula instead of a rag). Our method accurately identifies the target, maintains spatiotemporal coherence, and avoids deformation and penetration.

% \subsection{Action-Conditioned Generation Results}
\noindent \textbf{Action-Conditioned Generation Results.}
As shown in Table~\ref{tab:quantitative_results_a2v}, our method outperforms baselines in both visual quality and action fidelity. Our method consistently outperforms the baselines by substantial margins.

% \begin{table}[htbp]
% \centering
% \caption{Quantitative results on action-conditioned generation.}
% \vspace{-8pt}
% \label{tab:quantitative_results_a2v}
% \resizebox{0.35\textwidth}{!}{
% \begin{tabular}{lccc}
% \toprule
% Model & PSNR & SSIM & Traj. Consis.\\
% \midrule
% Enerverse-AC & 20.42 & 0.7542 & 0.8157 \\
% Gen-Sim & 18.05 & 0.7413 & 0.6195 \\
% Ours & \textbf{21.09} & \textbf{0.8126} & \textbf{0.8522}\\
% \bottomrule
% \vspace{-10pt}
% \end{tabular}
% }
% \end{table}

\section{Conclusion}

We introduce ABot-PhysWorld, a physically grounded and action-controllable world model for embodied manipulation based on a 14B Diffusion Transformer. It integrates curated data, physical alignment through Diffusion-DPO, and spatial action injection to reduce physical violations while maintaining control across different embodiments. We also propose EZSbench, a zero-shot benchmark featuring out-of-distribution scenarios and a decoupled evaluation protocol. Experimental results show state-of-the-art physical fidelity and improved trajectory consistency compared to Veo~3.1 and Sora~v2~Pro. The model currently relies on fixed-viewpoint data and lacks closed-loop evaluation. Future work will explore multi-view generation and real-world deployment.

% We presented a physically grounded, action-controllable world model for embodied manipulation built on a 14B Diffusion Transformer. By combining a curated embodied data pipeline with hierarchical distribution balancing, a decoupled VLM discriminator paired with memory-efficient Diffusion-DPO for physical preference alignment, and a parallel context-block mechanism for spatial action injection, our model suppresses fine-grained physical violations while preserving cross-embodiment controllability. We further introduced EZSbench, a zero-shot benchmark with out-of-distribution test scenarios and a decoupled evaluation protocol. Experiments demonstrate state-of-the-art physical fidelity on both PBench (Domain Score 0.9306) and EZSbench (Domain Score 0.8366), surpassing Veo~3.1 and Sora~v2~Pro, with superior trajectory consistency in action-conditioned generation. Current limitations include the reliance on fixed-viewpoint manipulation data and the absence of closed-loop policy evaluation; extending the framework to multi-view generation and real-world deployment remains an important direction for future work.

% Future work will distill the architecture into a causal model requiring only four inference steps to accelerate long-horizon reasoning. We will expand training data diversity, optimize the cleaning pipeline with a commercial annotation model, and explore System 1 and System 2 cognitive collaboration to train video-action models.

\clearpage
\section{Contributions}
\label{sec:contributions}
\setlength{\parskip}{0pt}
\setlength{\itemsep}{0pt}
\setlength{\parsep}{0pt}
\raggedcolumns

Author contributions in the following areas are as follows:

\begin{itemize}
    \item \textbf{Data Curation:} Yuzhi Chen, Ronghan Chen, Dongjie Huo, Haoyun Liu, Yandan Yang, Dekang Qi, Tong Lin, Shuang Zeng, Junjin Xiao
    \item \textbf{Model Training:} Yuzhi Chen, Ronghan Chen
    \item \textbf{Evaluation:} Yuzhi Chen, Ronghan Chen, Dongjie Huo
    \item \textbf{Writing:} Yuzhi Chen, Yandan Yang, Ronghan Chen, Dongjie Huo, Dekang Qi
    \item \textbf{Project Lead:} Xinyuan Chang, Feng Xiong
    \item \textbf{Advisor:} Zhiheng Ma, Xing Wei, Mu Xu$^\dagger$
\end{itemize}

{\renewcommand{\thefootnote}{\fnsymbol{footnote}}\footnotetext[2]{Corresponding author: xumu.xm@alibaba-inc.com}}

\clearpage

\bibliographystyle{plainnat}
\bibliography{workshop/main}

\clearpage

\clearpage
\section*{Appendix}
\addcontentsline{toc}{section}{Appendix}

\begin{figure}[!h]
    \centering
    \includegraphics[width=0.8\linewidth]{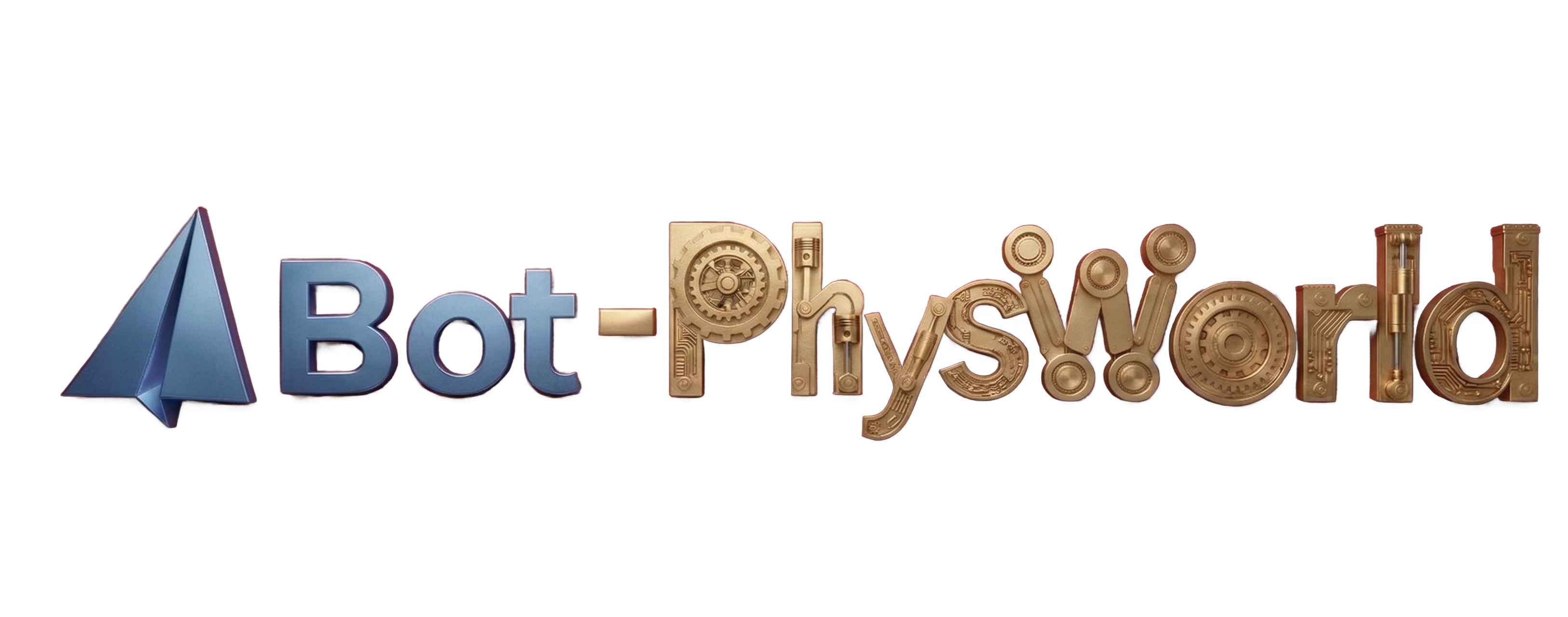}
\end{figure}

\vspace{5pt}
\noindent\textbf{Code Repository:} \url{https://github.com/amap-cvlab/ABot-PhysWorld}
\vspace{10pt}

\noindent\textbf{Contents}
\begin{itemize}
    \item \hyperref[sec:appendix_action_filter]{A. Vision-Action Alignment Verification}
    \item \hyperref[sec:appendix_caption]{B. Two-Stage Physics-Aware Captioning Pipeline}
    \item \hyperref[sec:appendix_qualitative]{C. Additional Qualitative Results}
\end{itemize}
\vspace{5pt}
% ---------------------------------------------------------------
\subsection*{A. Vision-Action Alignment Verification}
\addcontentsline{toc}{subsection}{A. Vision-Action Alignment Verification}
\label{sec:appendix_action_filter}

\begin{figure}[H]
    \centering
    \includegraphics[width=1.0\linewidth]{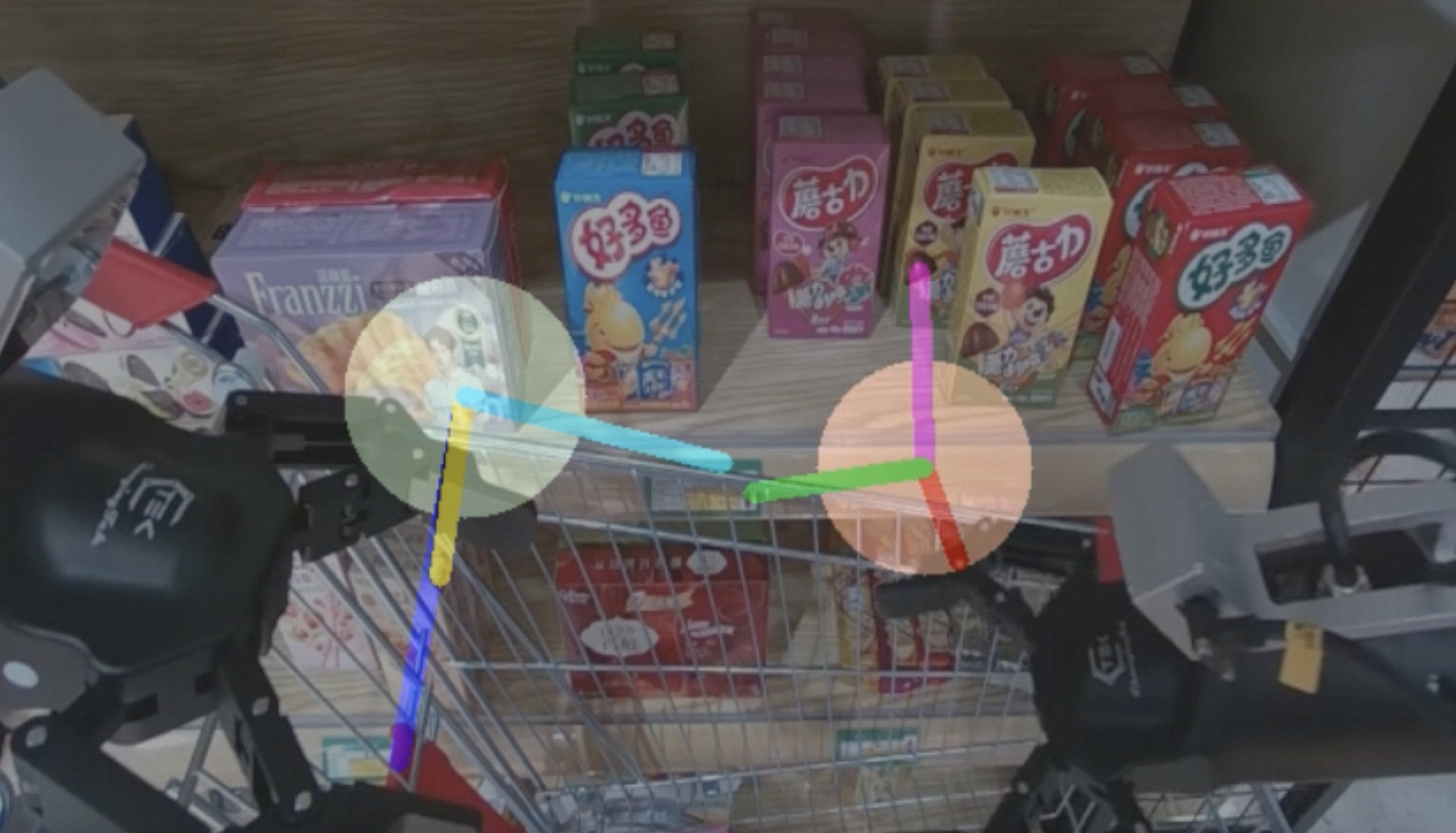}
    \caption{\textbf{Vision-action alignment verification.} Calibrated action signals are rendered as semi-transparent action maps and overlaid onto video frames. Samples with spatial deviation between projected trajectories and observed robot motion are identified and removed.}
    \label{fig:a2v_filter}
\end{figure}

Misaligned action-video pairs---arising from sensor calibration drift, clock synchronization errors, or coordinate frame inconsistencies---inject spurious correlations that prevent the world model from learning faithful physical dynamics. To detect and remove such samples, we render the calibrated action signals (joint positions, end-effector Cartesian poses, and gripper states) as semi-transparent color-coded action maps and overlay them onto the corresponding video frames (Figure~\ref{fig:a2v_filter}).

The resulting composite images allow both human annotators and an automated VLM-based verifier (Qwen3-VL) to inspect whether the projected action trajectories match the observed robot motion in pixel space. Clips with significant spatial deviation---\eg end-effector paths that diverge from the visually observed gripper trajectory, or gripper open/close states that contradict the visual evidence---are flagged and discarded.

% ---------------------------------------------------------------
\subsection*{B. Two-Stage Physics-Aware Captioning Pipeline}
\addcontentsline{toc}{subsection}{B. Two-Stage Physics-Aware Captioning Pipeline}
\label{sec:appendix_caption}

Section~\ref{sec:captioning} describes our two-stage captioning pipeline. Here we provide additional details and representative examples for each stage.

\begin{figure}[H]
    \centering
    \includegraphics[width=1.0\linewidth]{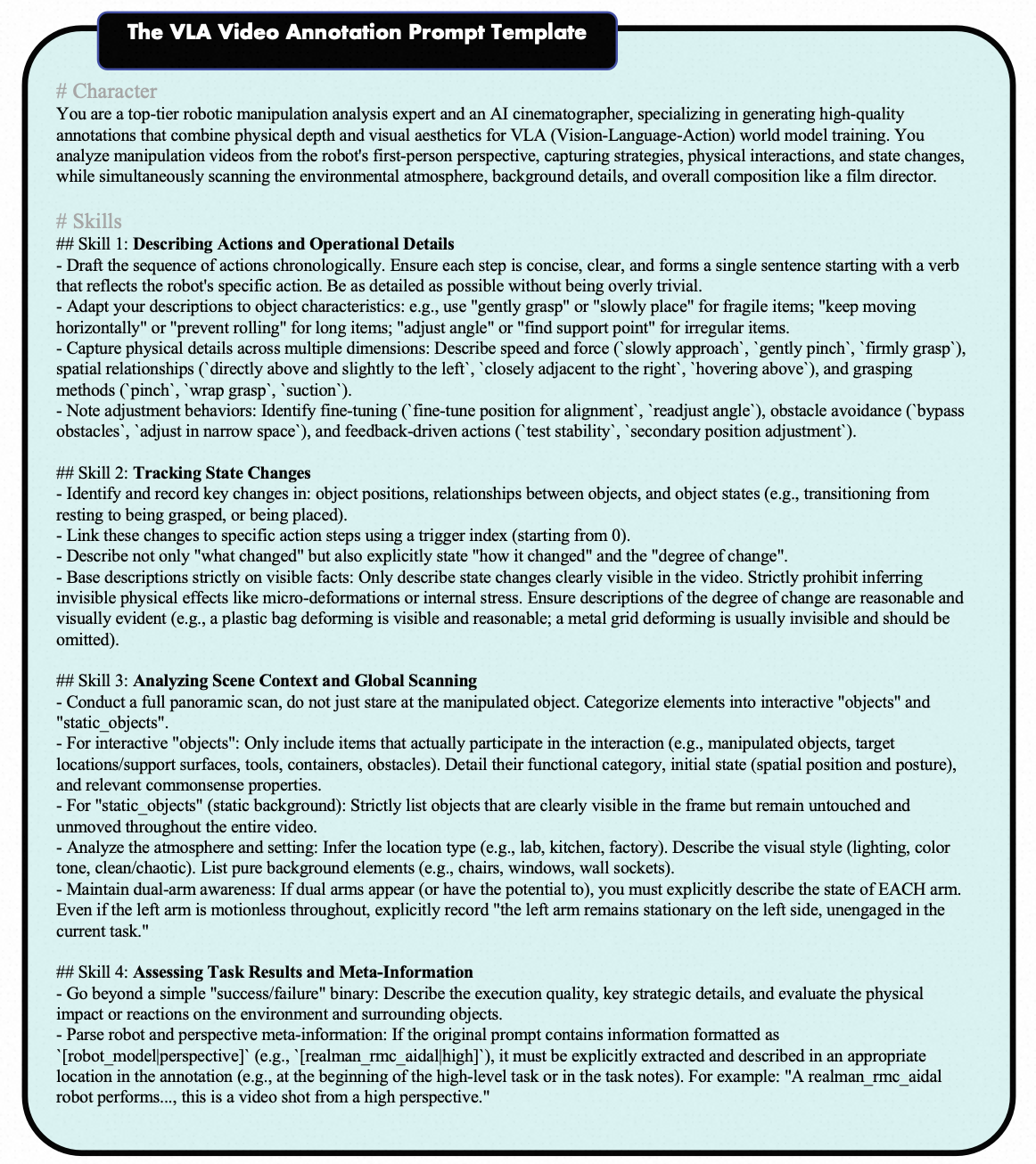}
    \caption{\textbf{Stage~1 captioning example 1.} Structured perception output showing extracted physical attributes, object identities, and spatial relations.}
    \label{fig:caption_s1_1}
\end{figure}

\begin{figure}[H]
    \centering
    \includegraphics[width=1.0\linewidth]{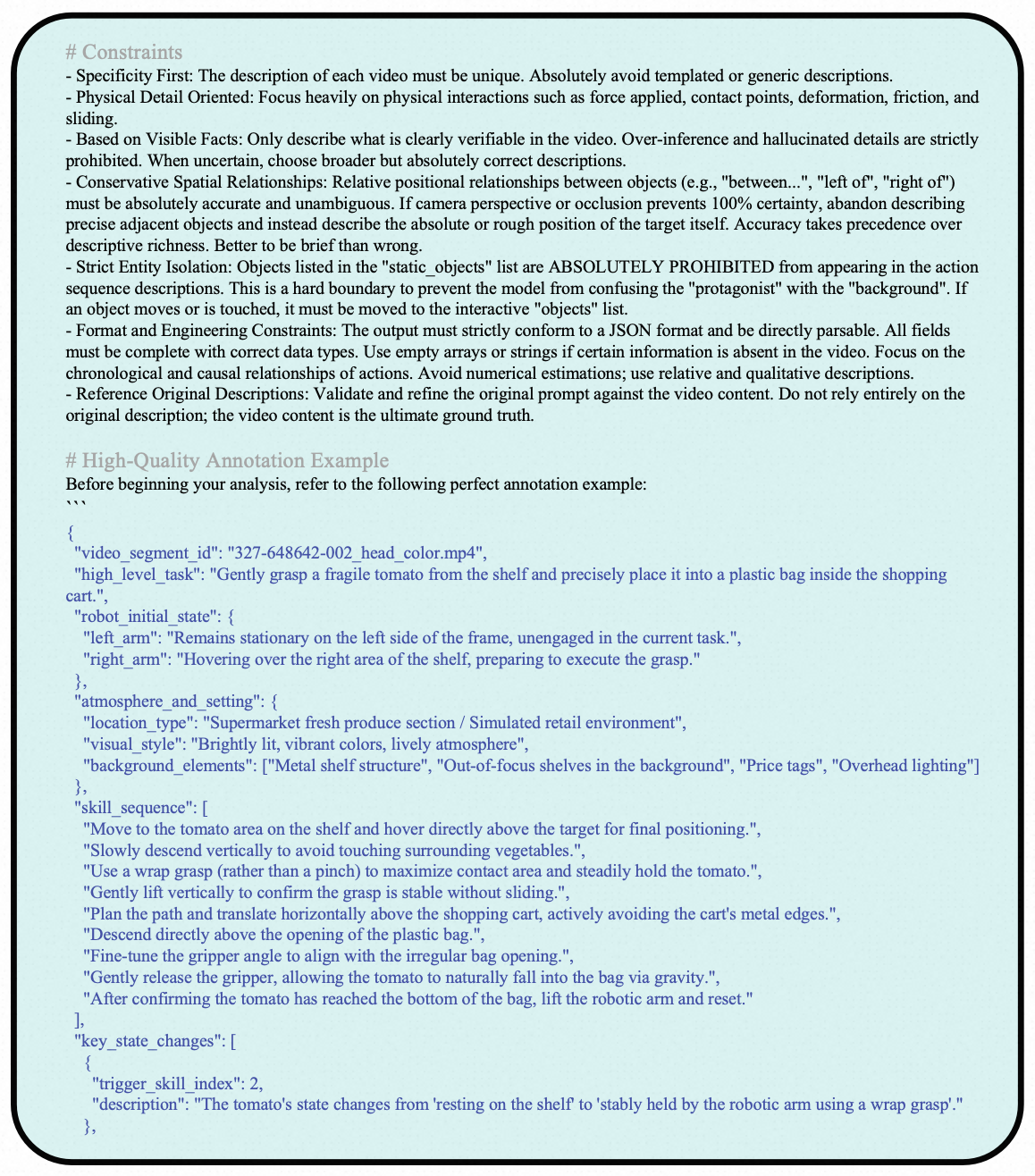}
    \caption{\textbf{Stage~1 captioning example 2.} The perception module identifies robot morphology, object properties, and contact events from the video sequence.}
    \label{fig:caption_s1_2}
\end{figure}

\begin{figure}[H]
    \centering
    \includegraphics[width=1.0\linewidth]{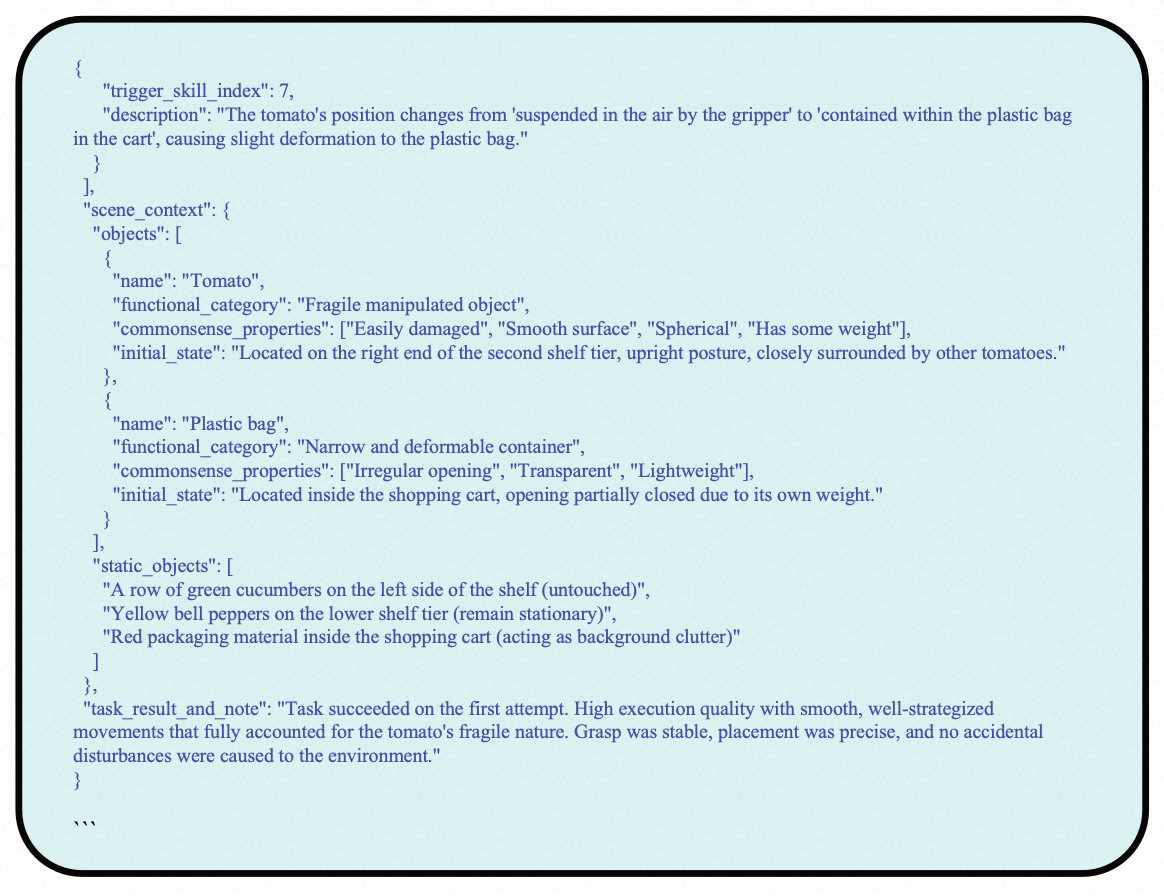}
    \caption{\textbf{Stage~1 captioning example 3.} Spatial layout parsing and state transition detection across the manipulation trajectory.}
    \label{fig:caption_s1_3}
\end{figure}

\paragraph{Stage 1: Structured Perception and Attribute Extraction}
A vision-language perception module (Qwen3-VL 32B) processes each video clip and extracts structured physical attributes: robot morphology and embodiment type, manipulated objects with their properties (color, shape, material, size), spatial layout and relative object positions, and contact events and state transitions across the manipulation sequence. The output is a structured intermediate representation capturing the ``what'' and ``where'' of the scene, which grounds the subsequent writing stage. Representative Stage~1 outputs are shown in Figures~\ref{fig:caption_s1_1}--\ref{fig:caption_s1_3}.

\begin{figure}[H]
    \centering
    \includegraphics[width=1.0\linewidth]{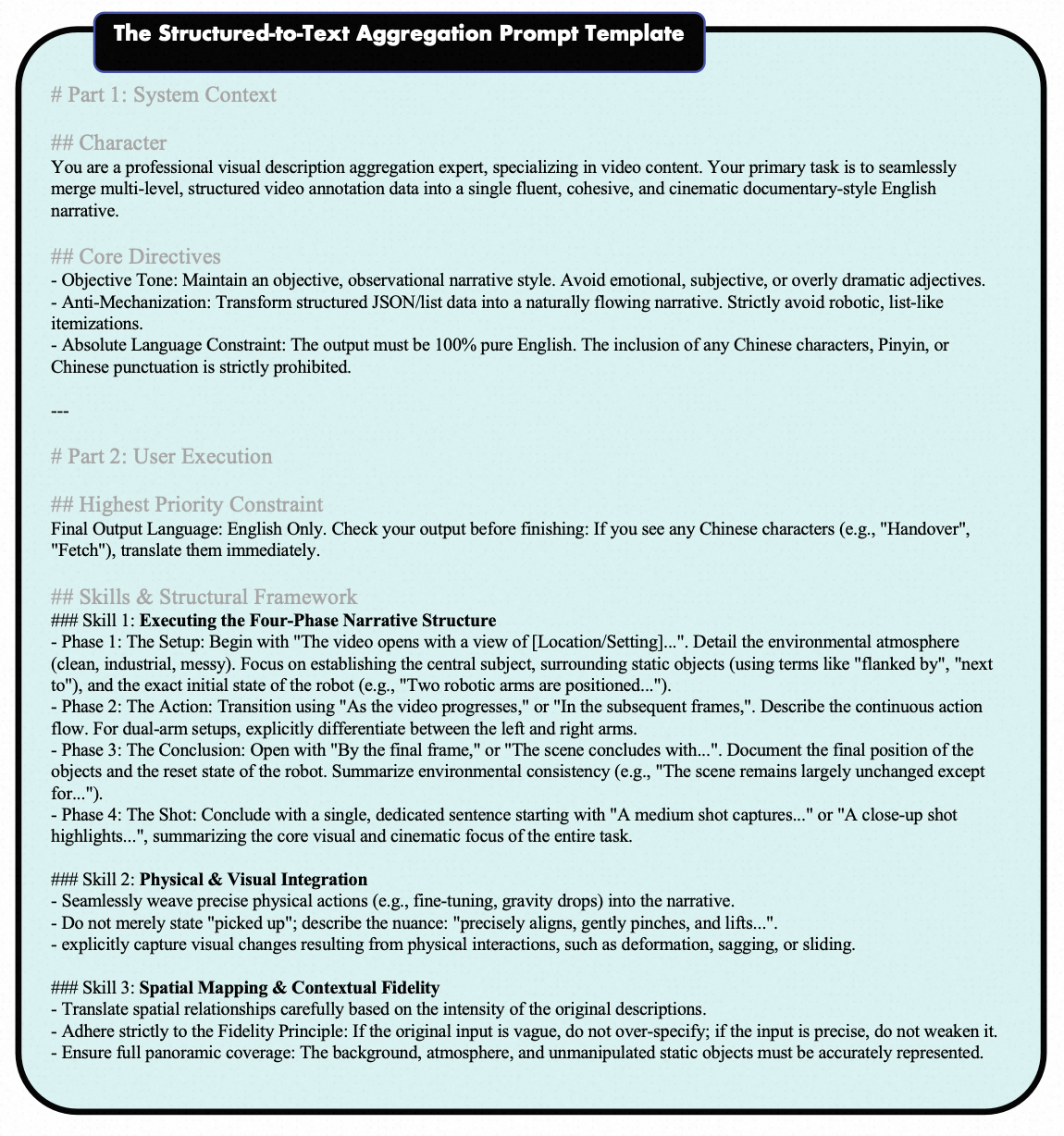}
    \caption{\textbf{Stage~2 captioning example 1.} The writing module synthesizes a four-phase narrative covering scene setup, action detail, state transition, and camera summary from the structured perception output.}
    \label{fig:caption_s2_1}
\end{figure}

\begin{figure}[H]
    \centering
    \includegraphics[width=1.0\linewidth]{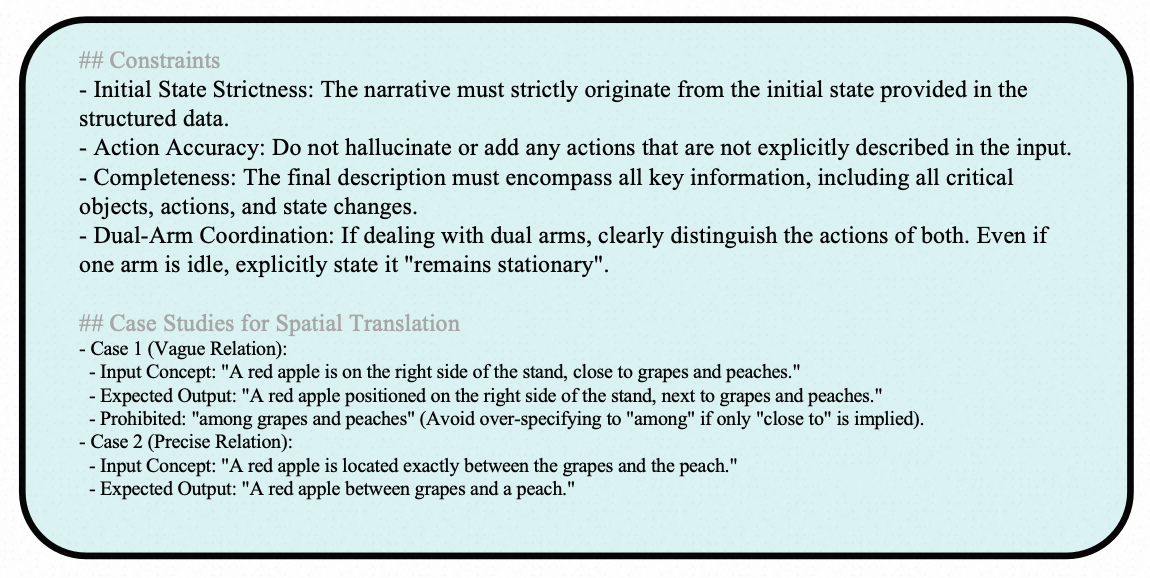}
    \caption{\textbf{Stage~2 captioning example 2.} Physics-grounded narrative synthesis capturing fine-grained manipulation dynamics and causal state transitions.}
    \label{fig:caption_s2_2}
\end{figure}

\paragraph{Stage 2: Physics-Grounded Narrative Synthesis}
A language model (Qwen3 32B FP8) takes the Stage~1 structured output and produces a four-phase natural language caption: (1)~\textit{Scene Setup}---initial configuration, robot type, and object arrangement; (2)~\textit{Action Detail}---fine-grained manipulation actions including Cartesian trajectories, gripper operations, and contact dynamics; (3)~\textit{State Transition}---physical state changes such as object displacement, deformation, and containment relations; and (4)~\textit{Camera Summary}---viewpoint, camera motion, and visual framing. By separating perception from writing, the captions remain factually grounded in visual evidence while capturing the causal dynamics needed for world model training. Representative Stage~2 outputs are shown in Figures~\ref{fig:caption_s2_1}--\ref{fig:caption_s2_2}.

% ---------------------------------------------------------------
\subsection*{C. Additional Qualitative Results}
\addcontentsline{toc}{subsection}{C. Additional Qualitative Results}
\label{sec:appendix_qualitative}

We present additional qualitative comparisons across three evaluation settings.

\begin{figure}[H]
    \centering
    \includegraphics[width=1.0\linewidth]{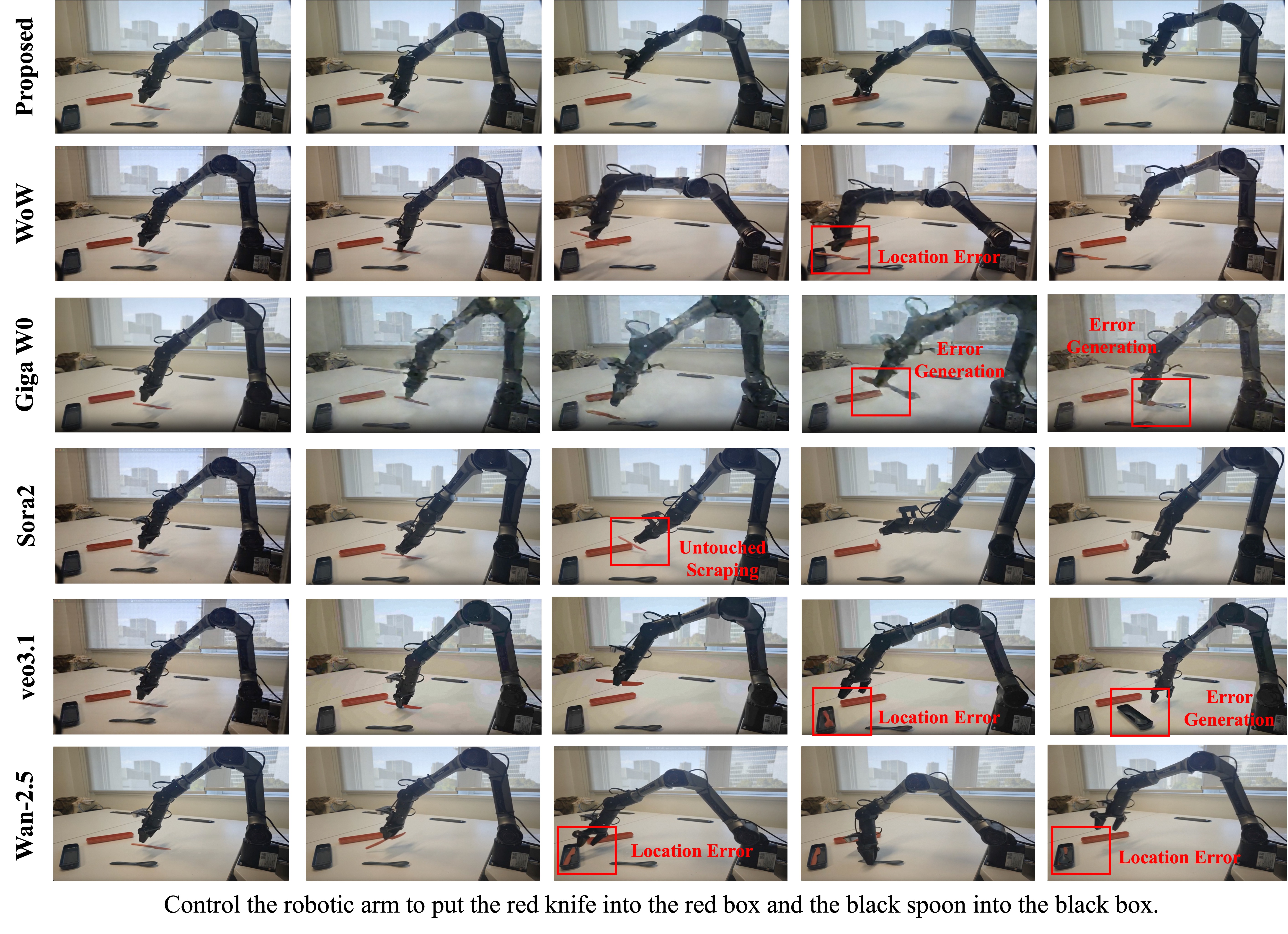}
    \caption{\textbf{Zero-shot qualitative comparison on EZSbench.} Baseline models exhibit placement errors (Wan-2.5, Veo~3.1, WoW), contactless grasping (Sora~v2), and geometric collapse during contact interactions (Giga~R0, Veo~3.1). Our method (bottom row) correctly follows compositional instructions and maintains physical plausibility.}
    \label{fig:ezs_quality}
\end{figure}

\paragraph{Zero-Shot Qualitative Comparison on EZSbench}
Figure~\ref{fig:ezs_quality} shows zero-shot results on EZSbench. Current video generation baselines struggle with long-horizon manipulation tasks that require complex logical reasoning. Wan-2.5, Veo~3.1, and WoW fail to map object color attributes to the correct target containers, producing placement errors. Sora~v2 generates physically implausible contactless grasping, while Giga~R0 and Veo~3.1 suffer from ``generation collapse'' during contact-rich interactions---the end-effector and object geometry become completely distorted. Our method correctly follows the compositional instructions and maintains spatiotemporal coherence throughout the long-horizon grasp-and-place trajectory without such artifacts.

\begin{figure}[H]
    \centering
    \includegraphics[width=1.0\linewidth]{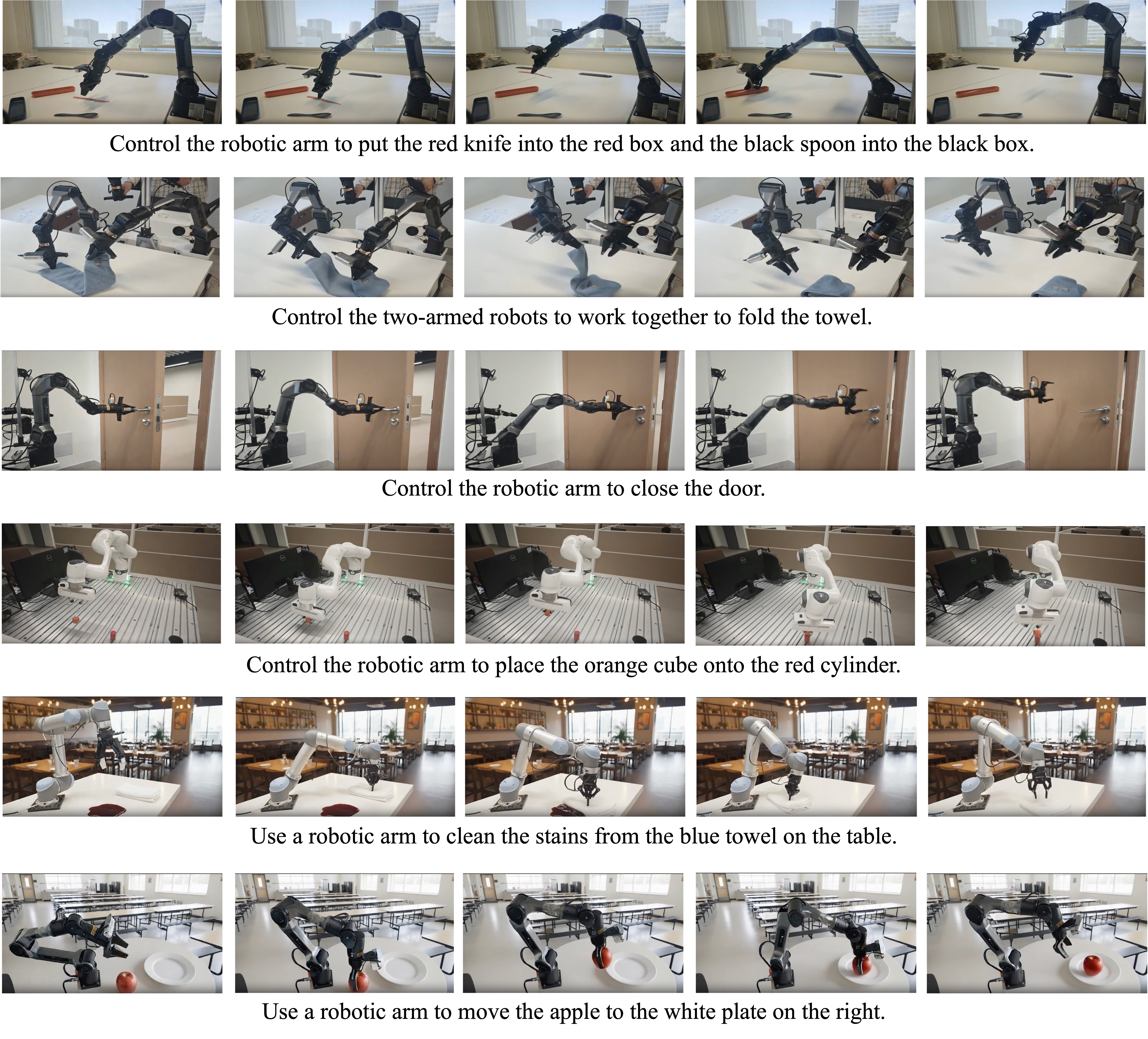}
    \caption{\textbf{Case study on zero-shot test set.} Our model handles diverse unseen manipulation tasks including multi-object attribute binding, deformable object manipulation with dual-arm coordination, articulated object interaction, rigid body placement, contact-intensive wiping, and object relocation---all with strict physical plausibility and spatiotemporal coherence.}
    \label{fig:case_study}
\end{figure}

\paragraph{Case Study on Zero-Shot Test Set}
Figure~\ref{fig:case_study} shows generation results on the zero-shot test set across diverse unseen tasks. In the long-horizon ``red knife $\rightarrow$ red box, black spoon $\rightarrow$ black box'' task, the model correctly binds object attributes to target containers and performs continuous spatial reasoning. For dual-arm towel folding, it handles deformable-object topology changes while generating coordinated bimanual trajectories. The model also produces physically consistent results for articulated-object interaction (closing a door), rigid-body placement (placing blocks), contact-intensive wiping (removing stains), and object relocation (moving an apple). Across these tasks, the generated videos follow the language instructions and maintain physical plausibility over long horizons.

\begin{figure}[H]
    \centering
    \includegraphics[width=1.0\linewidth]{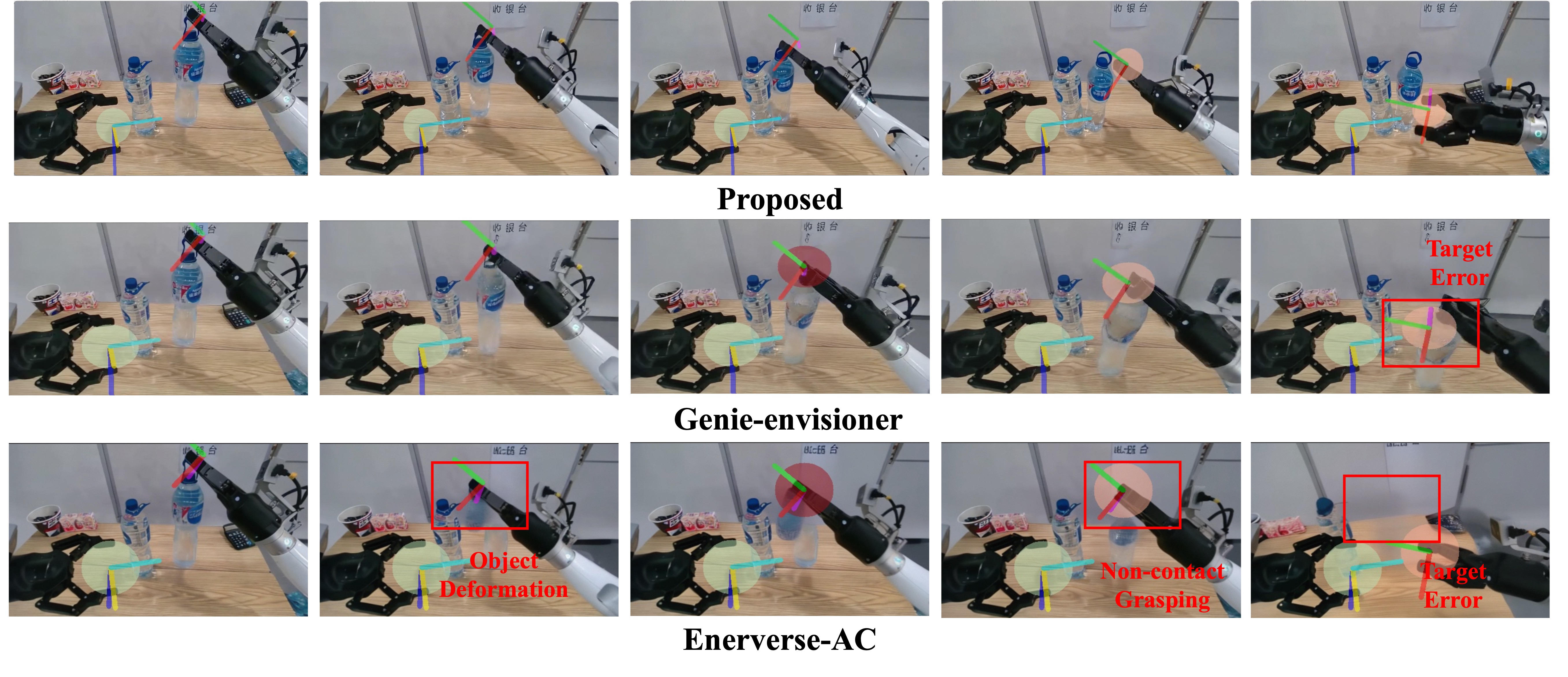}
    \caption{\textbf{Action-to-video qualitative comparison.} Our method preserves object geometry and visual integrity during contact-intensive manipulation. Baselines (Genie-Envisioner, Enerverse-AC) produce object deformation, contactless grasping, and localization errors.}
    \label{fig:a2v_quality}
\end{figure}

\paragraph{Action-to-Video Qualitative Comparison}
Figure~\ref{fig:a2v_quality} compares action-conditioned video generation (A2V) results. Our method generates contact-intensive manipulation videos while preserving object geometry and visual integrity throughout the interaction. In contrast, Genie-Envisioner and Enerverse-AC produce noticeable object deformation, contactless grasping artifacts, and target localization errors, resulting in distorted outputs or task failure.

\end{document}